\documentclass{article} 
\usepackage{colm2024_conference}

\usepackage{booktabs}
\usepackage{graphicx}
\usepackage{enumitem}
\usepackage{wrapfig}
\usepackage{algorithm}
\usepackage{algpseudocode}
\usepackage{natbib}
\usepackage{makecell}
\usepackage{booktabs}
\usepackage{array}
\usepackage{amsmath}
\usepackage{amsfonts}
\usepackage{threeparttable}
\usepackage{multirow}
\usepackage{verbatim}
\usepackage{caption}
\usepackage{longtable}
\usepackage{ragged2e}      
\usepackage{array}         
\usepackage{supertabular}
\usepackage{CJKutf8}
\usepackage{array} 
\usepackage[utf8]{inputenc} 
\usepackage[T1]{fontenc}
\usepackage[french,vietnamese,mongolian,greek,english]{babel}
\usepackage{pifont}
\usepackage{xcolor}
\usepackage{enumitem}
\usepackage{tablefootnote}
\usepackage{xspace}
\usepackage{textcomp}
\usepackage{makecell}
\usepackage{lscape} 
\usepackage{siunitx}
\usepackage{listings}
\usepackage{xcolor}
\usepackage{colortbl}
\definecolor{kuaishoublue}{HTML}{6D9EEB}
\hypersetup{
    colorlinks=true,   
}
\definecolor{dt}{gray}{0.7}
\newcommand{\cmark}{\ding{51}}%
\newcommand{\xmark}{\ding{55}}%
\usepackage{pifont}       
\usepackage{bbding}      
\usepackage{scrextend}

\usepackage{tgpagella}
\usepackage{latexsym}
\usepackage[T1]{fontenc}
\usepackage[utf8]{inputenc}
\usepackage{microtype}
\definecolor{mydarkblue}{rgb}{0,0.08,0.45}
\definecolor{citecolor}{HTML}{0071BC}
\usepackage{url}            
\usepackage{nicefrac}       
\usepackage{changepage}
\usepackage{xargs}          
\usepackage{wrapfig,lipsum,booktabs}
\usepackage{longtable}
\usepackage{subcaption}
\usepackage{endnotes}

\usepackage{fancyvrb}
\usepackage{fvextra}
\usepackage{pgfplots}
\usetikzlibrary{pgfplots.groupplots}
\pgfplotsset{compat=1.3}
\usepackage{tikz}
\usetikzlibrary{patterns}

\usepackage[most]{tcolorbox}
\definecolor{blue1}{HTML}{196ab1}
\definecolor{blue2}{HTML}{4886c1}
\definecolor{blue3}{HTML}{5e9bd6}
\definecolor{blue4}{HTML}{77b1e2}
\definecolor{blue5}{HTML}{bdd930}
\definecolor{blue6}{HTML}{dfebf6}

\definecolor{red1}{HTML}{de512c}
\definecolor{red2}{HTML}{f2642d}
\definecolor{red3}{HTML}{f68f58}
\definecolor{red4}{HTML}{febf92}
\definecolor{red5}{HTML}{f8e9c8}

\usepackage[capitalize,noabbrev]{cleveref}
\crefname{section}{Section}{\S\S}
\Crefname{section}{Section}{\S\S}
\crefname{table}{Table}{Tables}
\crefname{figure}{Figure}{Figures}
\crefname{algorithm}{Algorithm}{}
\crefname{equation}{eq.}{}
\crefname{appendix}{Appendix}{}
\crefformat{section}{Section #2#1#3}
\usepackage{multicol}
\usepackage{tcolorbox}

\usepackage{titlesec}
\titleformat*{\section}{\large\bfseries}

\title{Agentic-MME: What Agentic Capability Really Brings to Multimodal Intelligence?}

\title{Agentic-MME: What Agentic Capability Really Brings to Multimodal Intelligence?}

\author{
Qianshan Wei$^{1,2,*}$, Yishan Yang$^{3,*}$, Siyi Wang$^{3,*}$, Jinglin Chen$^{3}$, Binyu Wang$^{4}$,\\
Jiaming Wang$^{4}$, Shuang Chen$^{8}$, Zechen Li$^{3}$, Yang Shi$^{5}$, Yuqi Tang$^{6}$,\\
Weining Wang$^{1}$, Yi Yu$^{7}$, Chaoyou Fu$^{4}$, Qi Li$^{1,\dagger}$, Yi-Fan Zhang$^{1,\dagger}$\\
\footnotesize{
$^{1}$CASIA \quad
$^{2}$UCAS \quad
$^{3}$SEU \quad
$^{4}$NJU \quad
$^{5}$PKU \quad
$^{6}$BUAA \quad
$^{7}$NTU \quad
$^{8}$UCLA\\
$^{*}$Equal contribution \quad
$^{\dagger}$Corresponding Authors
}
}

\usepackage{tabularx}
\usepackage{amssymb}
\usepackage{mathtools}
\usepackage{amsthm}
\usepackage{bbm}
\usepackage{enumerate}
\usepackage{needspace}
\usepackage[normalem]{ulem}

\tcbuselibrary{skins,breakable}

\theoremstyle{plain}

\theoremstyle{definition}

\theoremstyle{remark}

\definecolor{ForestGreen}{RGB}{34,139,34}
\definecolor{myyellow}{RGB}{181,181,27}
\definecolor{mygreen}{HTML}{00897B}
\definecolor{myred}{HTML}{D32F2F}
\definecolor{headergray}{gray}{0.96}
\definecolor{rowhighlight}{HTML}{FFF3C7}
\definecolor{CadetBlue}{HTML}{5F9EA0}
\definecolor{customblue}{HTML}{4A90D9}

\definecolor{BlueGreen}{RGB}{13,152,186}
\definecolor{RedOrange}{RGB}{255,83,73}

\renewcommand{\cmark}{\textcolor{mygreen}{\ding{51}}}
\renewcommand{\xmark}{\textcolor{myred!70}{\ding{55}}}
\newcommand{\pmark}{\textcolor{orange}{\ding{108}}}

\begin{document}

\maketitle

\begin{abstract}
Multimodal Large Language Models (MLLMs) are rapidly evolving from passive observers into active agents. These agents increasingly solve problems through \textit{Visual Expansion}, actively invoking visual tools to transform images. Furthermore, as real-world tasks often require information beyond visual content, modern systems combine these visual operations with \textit{Knowledge Expansion} via open-web search. However, existing evaluations fall short in three critical aspects. (i) Lack flexibility and comprehensiveness in tool integration that supports heterogeneous tool interfaces. (ii) Test image tool use and web search separately, leaving their synergy unexplored. (iii) Evaluate primarily by the correctness of the final answer. Consequently, they cannot verify whether the tools were \textit{actually} invoked, applied \textit{correctly}, or used \textit{efficiently}.  To answer what agentic capability truly brings to multimodal intelligence, we introduce Agentic-MME, a process-verified benchmark for Multimodal Agentic Capabilities. Agentic-MME contains 418 real-world tasks across 6 domains and 3 difficulty levels designed to evaluate capability synergy, featuring over 2,000 stepwise checkpoints that average more than 10 person-hours of manual annotation per task. Each task is paired with (i) a unified evaluation framework that supports both sandboxed code execution and structured tool APIs, and (ii) a human reference trajectory annotated with stepwise checkpoints along dual-axis: \textit{S-axis} and \textit{V-axis}. To enable true process-level verification, we move beyond final answers by auditing fine-grained intermediate states. We also quantify efficiency via an \textit{overthinking} metric relative to human trajectories. Experimental results show that the best model Gemini3-pro achieves an overall accuracy of 56.3\% and this score falls significantly to \textbf{23.0\%} on Level-3 tasks, underscoring the difficulty of real-world multimodal agentic problem solving.\textbf{Keywords:} Agentic MLLMs, Benchmark, Deep-Research
\end{abstract}


\begin{figure}[ht]
  \centering
  \includegraphics[width=0.87\linewidth]{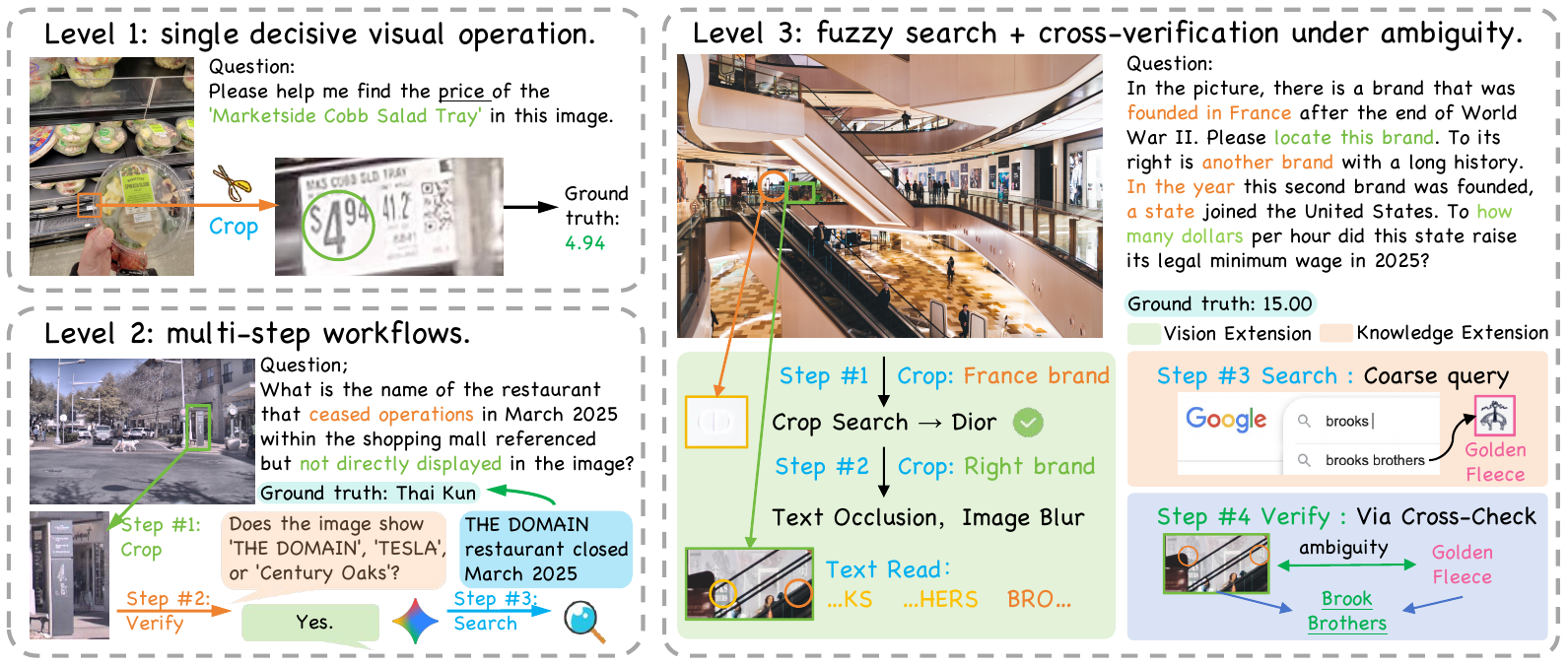}
  \caption{Case Studies in Agentic-MME across three difficulty levels. The examples highlight the benchmark's escalating complexity, evolving from isolated visual operations in Level 1 to deeply synergistic, multi-round visual and knowledge workflows in Level 3.}
  \label{fig:case}
\end{figure}

\newpage
{
  \setstretch{0.7}
  \tableofcontents
  \noindent\hrulefill
}
\newpage

\section{Introduction}
\label{sec:intro}

Multimodal Large Language Models (MLLMs) are rapidly evolving from passive observers to active investigators~\citep{shen2023hugginggpt,deng2023mind2web,mavors,zhang2025skywork,hong2025deepeyesv2}. Rather than answering from a static snapshot, modern systems increasingly solve problems by \emph{interacting}: they manipulate images to surface fine-grained evidence and consult external resources to verify facts that are not visually present. We refer to this shift as \emph{multimodal agentic capability}, which can be decomposed into two core dimensions: \textbf{(1) Visual Expansion}, which enables models to \textit{think with images} by actively transforming and analyzing inputs (e.g., cropping, rotating, enhancing) to uncover latent cues~\citep{wei2026zooming,su2025thinking,lai2025mini,zhang2025thyme,zheng2025deepeyes,monet}; and \textbf{(2) Knowledge Expansion}, which enables models to go beyond parametric memory via open-ended web search to validate real-world facts and resolve ambiguity~\citep{chen2025mindwatcher,zeng2026vision,narayan2025deepmmsearch}. While this active paradigm promises to solve complex real-world problems, current evaluation for \emph{multimodal agentic capabilities} remains fragmented and insufficient.

Most existing benchmarks capture specific aspects of tool use but fall short in three critical dimensions: \textbf{First, a lack of flexibility and comprehensiveness in tool integration.} Current evaluations often decouple visual tool use from open-web search, treating them as independent modules. Most benchmarks do not provide a unified framework that allows an agent to fluidly select and switch between arbitrary visual and search tools~\citep{li2025tir,guo2025beyond,li2025mm,tao2025mmsearch,froger2026gaia2,zhang2026adaptmmbench}, which severely restricts the upper bound of agentic behavior. \textbf{Second, the synergy between visual and knowledge expansion remains largely unexplored.} There are few realistic scenarios designed to test whether these two capabilities can work in concert. A true multimodal agent should excel in "intertwined" tasks that cannot be solved by simple visual expansion or isolatedknowledge expansiion alone—it is in these deeply coupled workflows that the unique value of an agencitc MLLM is realized. \textbf{Finally, a lack of rigorous process verification.} Existing evaluations focus primarily on final-answer correctness, offering limited insight into whether tools were actually invoked, applied correctly, or used efficiently~\citep{hou2025codev}. When evaluation only inspects the final response, it becomes impossible to diagnose whether a failure stems from limited perception, skipped tool calls, incorrect execution (e.g., cropping the wrong region), or redundant trial-and-error. Without stepwise grounding, these failure modes are conflated, obscuring the true bottlenecks of multimodal intelligence.

To bridge these gaps, we introduce Agentic-MME, a process-verified benchmark for \emph{multimodal agentic capabilities} through a unified and synergistic lens. Agentic-MME targets a realistic agent workflow: an agent must actively manipulate images to extract localized cues and, when necessary, coordinate these cues with open-web retrieval to obtain missing knowledge. 
To systematically evaluate these capabilities, we stratify tasks into three progressive difficulty levels based on \emph{interaction complexity} (Section~\ref{sec:task_setup}):
\begin{itemize}[leftmargin=*]
\vspace{-2mm}
\item \textbf{Level 1 (Visual Expansion Focus):} Tasks isolated to a single visual operation (e.g., one crop), testing the fundamental perception-action loop.
\item \textbf{Level 2 (Level 1+Knowledge Expansion):} Tasks requiring a simple combination of visual and knowledge expansion. Agents typically extract a visual cue and perform a web search to verify the missing fact. These tasks are solvable through simple tool-chaining, usually within three interaction rounds.
\item \textbf{Level 3 (Synergistic Coupling):} Challenging, realistic scenarios that demand iterative, interleaved execution of both visual and search tools. In these cases, agents extract tentative visual hints, use them to query the web for validation, and retrieve knowledge to further guide refined visual operations.
\end{itemize}
The design of Level 3 cases also represents our preliminary exploration into the synergy between visual and knowledge expansion. For instance, to identify an obscure logo (as shown in Figure~\ref{fig:case}), a model first perform a crop to surface blurry visual hints, then execute a multi-hop search to match potential brands for cross-validating. Neither using only visual tools nor a blind web search can solve such tasks in isolation. In this way, Agentic-MME provides a rigorous diagnostic tool to determine whether an agent fails at individual expansion capabilities or at the complex coordination required for high-order multimodal reasoning.

To address the outcome-based evaluation gap and enable true process verification, Agentic-MME moves beyond final-answer grading by fully human-annotated checkpoints and reference trajectories. This design requires substantial \textbf{high-quality human effort}: Agentic-MME features over 2000 stepwise checkpoints, averaging {10+ person-hours} of careful manual annotation and verification per task, to make intermediate tool use and evidence explicitly verifiable.
We evaluate intermediate behavior along two complementary axes. The \textbf{S-axis} audits \textit{Knowledge Expansion} by evaluating search-related strategies; for every search hop, we associate checkpoints with human-verified keywords, reference URLs, and expected intermediate answers.
The \textbf{V-axis} audits \textit{Visual Expansion} by checking both the proactive intent to invoke tools and whether intermediate artifacts genuinely contain the required visual cues. Concretely, each V-axis checkpoint is paired with a targeted question and ground-truth intermediate processed images.

\begin{table}[t]
\centering
\caption{Comparison with representative benchmarks. 
\textbf{ImgTools}: dedicated visual manipulation tools;
\textbf{SearchCore}: open-web search is the core task;
\textbf{ProcessVerified}: explicit process verification;
\textbf{Unified}: protocol supports both code \& function-calling for visual tools;
\textbf{Efficiency}: cost/efficiency metric;
\textbf{Levels}: explicit difficulty levels splits.
(\pmark: partly supported.)}
\vspace{-3mm}
\label{tab:compare}
\aboverulesep=0pt 
\belowrulesep=0pt 
\renewcommand{\arraystretch}{1.1} 
\setlength{\tabcolsep}{4pt}      

\definecolor{zebra}{gray}{0.96} 
\definecolor{rowhighlight}{HTML}{FFF3C7} 

\resizebox{\columnwidth}{!}{%
\begin{tabular}{lcccccc}
\toprule
\multirow{2.5}{*}{Benchmark} 
& \multicolumn{2}{c}{Capabilities} 
& \multicolumn{3}{c}{Evaluation Protocol} 
& \multirow{2.5}{*}{Levels} \\

\cmidrule(lr){2-3} \cmidrule(lr){4-6}

& \makecell{Img\\Tools} 
& \makecell{Search\\Core} 
& \makecell{Process\\Verified} 
& \makecell{Unified\\Code+Tool} 
& \makecell{Efficiency\\Metric} 
& \\
\midrule


GTA \citep{wang2024gta}              & \cmark & \xmark & \xmark & \xmark & \xmark & \xmark \\

\rowcolor{zebra}
m\&m's \citep{ma2024m}               & \cmark & \xmark & \xmark & \xmark & \xmark & \xmark \\

TIR-Bench \citep{li2025tir}          & \cmark & \xmark & \xmark & \xmark & \xmark & \xmark \\

\rowcolor{zebra}
VisToolBench \citep{guo2025beyond}   & \cmark & \xmark & \xmark & \xmark & \xmark & \xmark \\

MMSearch \citep{jiang2024mmsearch}   & \xmark & \cmark & \xmark & \xmark & \xmark & \xmark \\

\rowcolor{zebra}
MMSearch-Plus \citep{tao2025mmsearch}& \pmark & \cmark & \xmark & \xmark & \xmark & \cmark \\

MM-BrowseComp \citep{li2025mm}       & \xmark & \cmark & \cmark & \xmark & \xmark & \cmark \\

\rowcolor{zebra}
GAIA2 \citep{froger2026gaia2}        & \xmark & \pmark & \cmark & \xmark & \cmark & \xmark \\

AdaptMMBench \citep{zhang2026adaptmmbench} & \cmark & \xmark & \cmark & \xmark & \pmark & \cmark \\
\midrule

\rowcolor{rowhighlight} 
\textbf{Agentic-MME}                     & \cmark & \cmark & \cmark & \cmark & \cmark & \cmark \\
\bottomrule
\end{tabular}%
}
\vspace{-4mm}
\end{table}

To ensure fair and comparable evaluation across heterogeneous implementations, Agentic-MME provides a standardized execution harness that unifies \textit{sandboxed code execution} and \textit{function-calling tool APIs}.
This unification is non-trivial in practice: different models exhibit drastically different code styles, and their sandboxes vary in libraries, I/O conventions, and artifact formats, which can introduce spurious advantages if left uncontrolled. We perform \textbf{model-specific, fine-grained environment engineering} and \textbf{normalization} so that outputs are produced, logged, and scored under a consistent protoco. We additionally conduct \textbf{human calibration and correction} of execution flows and artifact parsing to eliminate unfairness. For code-writing agents, we implement an AST-based tracer that extracts canonical visual operations from executed code traces, enabling consistent scoring across heterogeneous coding patterns and interaction modes.
Finally, we measure efficiency by comparing agent behavior against human reference trajectories and penalizing redundant tool calls via an \textit{overthinking} metric. In summary, our contributions are as follows:
\begin{itemize}[leftmargin=*,itemsep=-0.05em,topsep=0.2em]
\item[\ding{182}] \textbf{\textit{Agentic-MME Benchmark.}}
We introduce Agentic-MME, a benchmark that evaluates comprehensive \textbf{multimodal agentic capabilities}, which integrates \textit{visual tool use} with open-ended \textit{web search} in one workflow, with 418 real-world tasks across 3 difficulty levels, and 6 domains.
\item[\ding{183}] \textbf{\textit{Process-Verified Stepwise Evaluation.}}
We provide human reference trajectories with \textbf{2,000} stepwise checkpoints for grounding intermediate behavior: (i) the \textbf{V-axis} audits \textit{Visual Expansion} by verifying both the correctness of proactive intent to invoke tools and the faithfulness of intermediate visual artifacts; (ii) the \textbf{S-axis} audits \textit{Knowledge Expansion} by verifying search keywords and the correctness of retrieved intermediate answers. The same stepwise signals also capture redundant tool usage relative to human references.
\item[\ding{184}] \textbf{\textit{Unified Evaluation Framework Across Code and Tool Calls.}}
We develop a standardized execution and evaluation harness supporting both sandboxed Python code and structured function-calling APIs. We introduce an \textbf{AST-based tracer} that audits executed visual operations in code mode for consistent scoring across interaction styles.
\item[\ding{185}] \textbf{Empirical study and insights.} We evaluate leading proprietary and open-source models. Our findings expose a fundamental gap: while frontier models possess vast knowledge, they still struggle severely with reliable multi-step planning and precise tool-execution in real-world workflows.
\end{itemize}

\section{Agentic-MME}
\vspace{-2mm}
\subsection{overview}
\label{sec:benchmark_overview}
We introduce Agentic-MME, which is designed to evaluate multimodal agentic capabilities. Unlike existing works that isolate visual operations, Agentic-MME targets a realistic scenario where agents actively utilize visual tools to transform and perceive image content, and, contingent on task requirements, coordinate with open-web search to retrieve essential external knowledge.
\subsection{Task Setup, Difficulty, and Metrics}
\label{sec:task_setup}
\noindent\textbf{Task setup.}
Each instance provides one or more images and a question. Agents solve the task by actively manipulating the image within a unified tool-augmented interface, which equips with 13 distinct visual operations for \textit{Visual Expansion} and 4 open-web retrieval tools for \textit{Knowledge Expansion}. Implementation details are in Appendix~\ref{app:tools} and Sec.~\ref{sec:harness}.

\vspace{1mm}
\noindent\textbf{Task Scenarios and Difficulty levels.}
Agentic-MME encompasses two core scenarios: extracting hidden evidence through active image manipulation, and utilizing visual cues to retrieve or verify facts from the open web. We systematically stratify these tasks into three difficulty levels based on the \emph{interaction complexity} along a reasonable solution path: 
\begin{itemize}[leftmargin=*,itemsep=-0.1em,topsep=0.1em]
  \renewcommand\labelitemi{$\diamond$}  
  \item Level~1 requires a single visual operation (e.g., one \textit{crop}/ \textit{rotate}). 
  \item Level~2 requires multi-step workflows where agents actively manipulate the image to obtain key visual cues and may leverage open-web to retrieve external factual knowledge beyond visual content. (e.g., \textit{crop}$\rightarrow$ \textit{rotate} $\rightarrow$ \textit{ web-search} ). \item Level~3 represents advanced synergistic workflows where the answer cannot be reached through simple sequential tool chaining. Instead, it demands an intertwined, multi-round interaction between visual manipulation and web search. This level encompasses (i) integrating \emph{multiple clues} scattered across different regions or separate images, (ii) performing \emph{advanced CV analysis} (e.g., frequency-domain transformations for recognition), and (iii) resolving severe visual ambiguity (e.g., Figure~\ref{fig:case}). For the latter, agents must execute an explicit \textit{hypothesis--verification loop}. For example, to identify a highly blurry logo, an agent must first crop the image to extract a vague visual hint, perform multi-hop searches to retrieve candidate entities, and then synergistically cross-validate these retrieved external facts against the processed image to confirm the uncertain information.\end{itemize} This continuous cross-referencing between the visual input and the open web represents the true synergy required for real-world agentic problem-solving. More case studies are in Appendix~\ref{app:cases}.

\vspace{1mm}
\noindent\textbf{Metrics.}
We report final answer accuracy (\textbf{Acc}) together with process-aware and efficiency metrics derived from stepwise checkpoints.
Concretely, \textbf{S} and \textbf{V} are computed as the fraction of passed checkpoints on the S-axis and the V-axis. Because simply invoking a visual tool does not guarantee that the correct visual evidence is successfully extracted, we decouple the V-axis for finer diagnosis. We split V-axis into: \textbf{$V_{tool}$} : whether the agent initiates the required visual tool use at the right step; and \textbf{$V_{true}$} : whether the generated true intermediate visual artifact actually contain the require evidence. To evaluate efficiency, we introduce an \textbf{Overthink} metric to penalize redundant behaviors relative to the human minimal trajectory, defined as $\text{Overthink}=\max(0, C_{\text{agent}}-C_{\text{human}})/(C_{\text{human}}+1)$, where $C$ counts the interactions for tasks. Exact scoring rules (checkpoint matching, evidence verification) are specified in Appendix.~\ref{app:eval_protocol}.

\subsection{Data Collection and Annotation}
\label{sec:collection}
\begin{figure}[t]
  \centering
  \includegraphics[width=\linewidth]{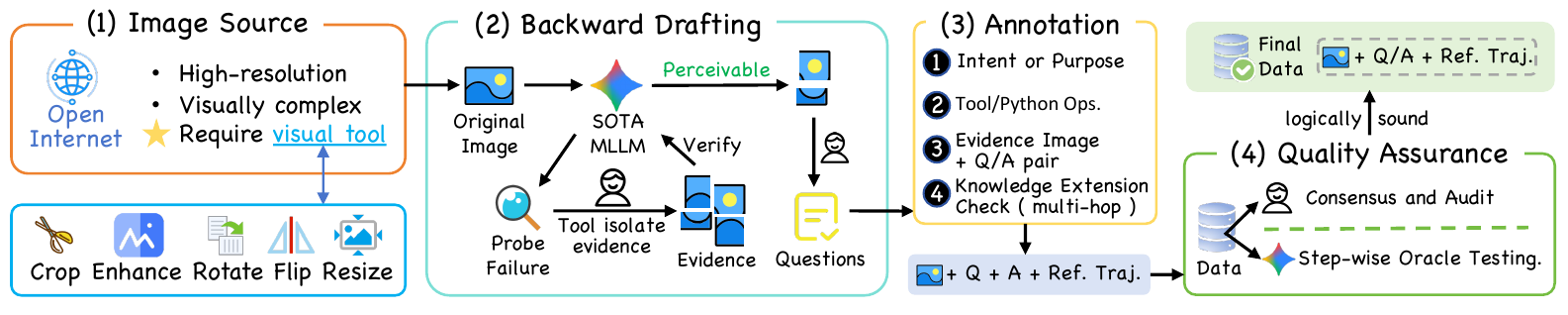}
  \vspace{-6mm}
  \caption{Data Collection and Annotation pipeline, including image sourcing, backward drafting, granular step-wise annotation, and quality assurance.}
  \label{fig:anno_pipeline}
  \vspace{-4mm}
\end{figure}
Process-level supervision is only meaningful when annotations are rigorous, reproducible, and verifiable. We formulate data construction as a strictly controlled protocol  (as shown in Fig~\ref{fig:anno_pipeline}).

\vspace{1mm}
\noindent\textbf{Image Sourcing.}
We source high-resolution, visually complex images across 6 heterogeneous domains from the open web. To prevent the benchmark from degrading into a simple cropping task, annotators intentionally curate real-world scenarios that require a diverse visual toolkit. Specifically, the dataset includes oversized images requiring localized \emph{cropping}, low-light environments or folded documents needing photometric \emph{enhancement}, misaligned documents requiring \emph{rotation}, mirrored visuals demanding \emph{flipping} and steganographic visuals requiring \emph{resizing} . By embedding decisive evidence within these challenging contexts, the dataset strictly mandates active and diverse visual manipulation.

\noindent\textbf{Model-in-the-Loop Backward Drafting.}
Contributors must pass qualification tasks to align on difficulty and granularity. To rigorously guarantee that visual interaction is necessary, tasks are designed via an empirical \textit{model-in-the-loop backward-drafting} mechanism. Annotators first prompt a state-of-the-art model (e.g., Gemini 3 Pro) to comprehensively describe the raw image. By reviewing this initial response, annotators specifically target visual details that the model overlooks or hallucinates during passive inspection. The annotator then applies visual manipulation tools (e.g., cropping) to isolate the evidence, and verifies the \emph{same model} can successfully perceive the processed image. Crucially, the model's updated response is cross-checked against a human ground truth to ensure it is genuinely correct rather than a confident hallucination. Once validated, the annotator drafts the user question targeting this hidden evidence, making active visual manipulation a prerequisite.

\vspace{1mm}
\noindent\textbf{Granular Step-wise Annotation.}
Beyond only supervise the final answer, we meticulously annotate the full trajectory. For \emph{every single step} in the reference trajectory, annotators explicitly record: 
(1) a precise natural-language \textbf{description} detailing the exact intent and purpose of the action; 
(2) the specific tool or Python operations required. To support robust automated evaluation, for \emph{each} visual checkpoint, annotators establish the structural ground-truth necessary for our AST-based parsing; (3) for steps yielding new processed images, annotators record the corresponding intermediate visual artifacts (e.g., groundtruth crops), paired with a highly specific intermediate test question (e.g., "What is the road name shown in the image?") and its expected answer. This ensures that the processed image is objectively verifiable by an external judge; and 
(4) for external knowledge, annotators meticulously document \emph{every single hop} of complex multi-hop searches, recording the required \textit{search keywords}, \textit{human-verified URLs}, and the \textit{expected answer} necessary to the task.

\vspace{1mm}
\noindent\textbf{Answer Standardization.}
To eliminate evaluation ambiguity and avoid reliance on LLM-as-a-judge for final accuracy, we proactively constrain the output space during task design. Annotators specifically draft questions that inherently yield concise, verifiable answers. Furthermore, each instance includes explicit prompt instructions dictating the required final format (e.g., specific units, categorical options, or short strings). Supported by an accepted-variant list for minor deviations, this prompt-level control ensures that final answer extraction and scoring remain deterministic and regex-friendly.
\begin{figure}[t]
  \centering
  \includegraphics[width=\linewidth]{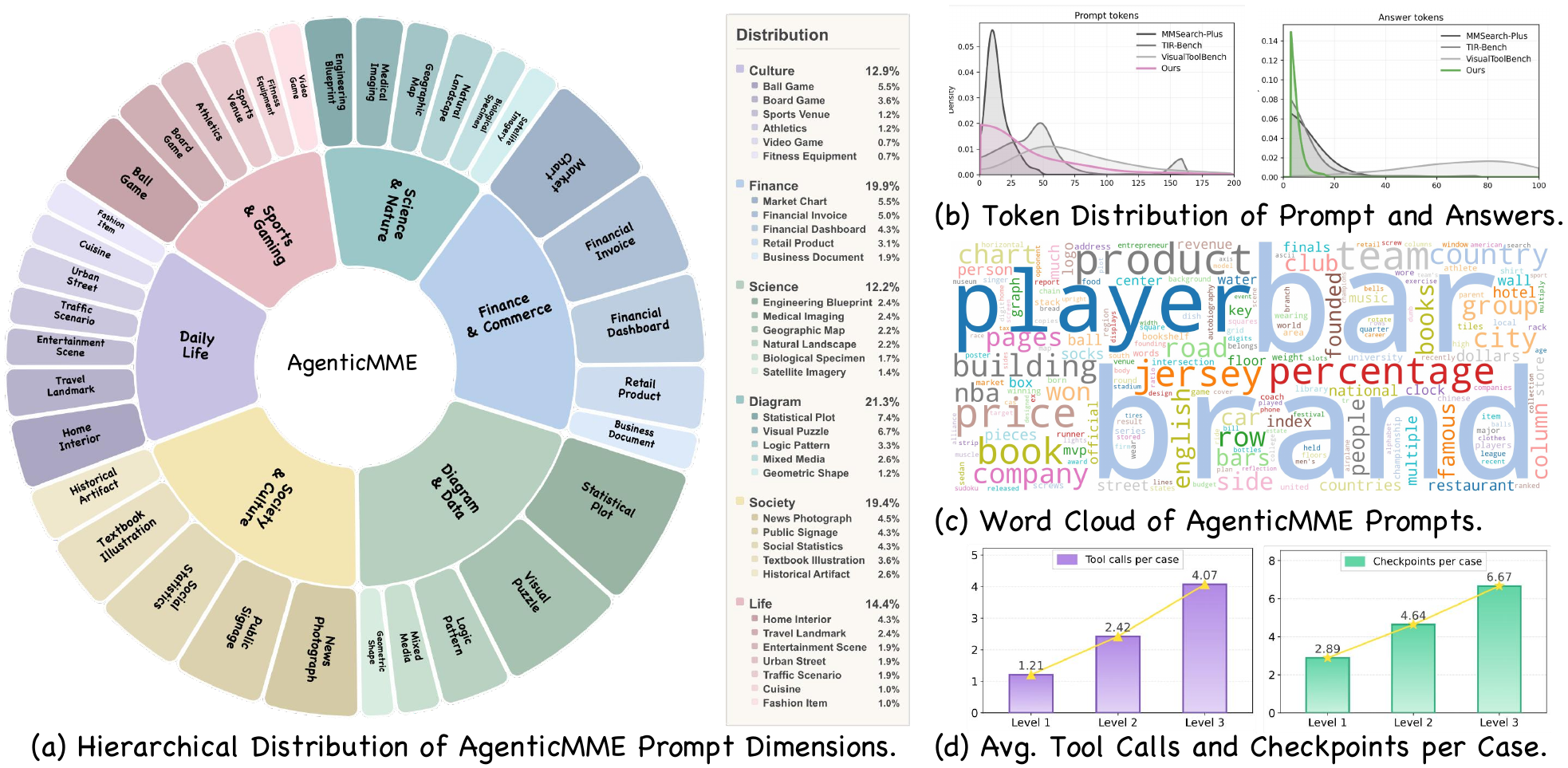}
  \vspace{-6mm}
  \caption{Overview of Agentic-MME Dataset Statistics. The benchmark exhibits broad domain and semantic diversity, with increasing tool calls and checkpoints reflecting the escalating demand for long-horizon reasoning across difficulty levels.}
  \label{fig:data}
  \vspace{-4mm}
\end{figure}

\subsection{Quality Control and Assurance}
\label{sec:qa}
To guarantee that tasks are logically sound, every drafted instance undergoes rigorous human-model co-verification.

\vspace{1mm}
\noindent\textbf{Human Verification: Consensus and Audit.}
To guarantee human solvability and annotation quality, each drafted instance is evaluated by the author and at least two independent verifiers. Verifiers first attempt the task end-to-end without accessing the reference path. Given the visual complexity, they may use manual tools (e.g., web browsers, image viewers) and tool-less vision-language models strictly for perceptual assistance. While the vast majority of instances are successfully solved on the first attempt, any divergent answer triggers a joint review. The team carefully refines ambiguous questions, permanently discarding any instances with unresolved disagreements to maintain the rigor of the dataset. Once strict consensus is reached on the final answer, verifiers systematically audit step-wise annotations, ensuring that all intermediate tool actions, visual artifacts, and retrieved URLs are accurate and logically sound.

\vspace{1mm}
\noindent\textbf{Model Verification: Step-wise Oracle Testing.}
To ensure that current models are fundamentally capable of solving the tasks when following the correct trajectory, we probe all instances using state-of-the-art models (e.g., Gemini 3 Pro). Specifically, we perform \textit{step-wise oracle guided verification}: we walk the model through the human reference path, providing the exact ground truth intermediate observations (e.g., the properly cropped image region or the correct web text) at key steps. A task is retained \emph{if and only if} the model can successfully arrive at the correct final answer. This check confirms that the underlying visual and textual evidence is perceptible to the model. Consequently, an agent's failure during evaluation directly reflects shortcomings in tool execution or multi-step planning, rather than a inability to process the evidence.

\vspace{1mm}
\noindent\textbf{Dataset Statistics.}
\definecolor{colorNoTool}{HTML}{B2DFDB} 
\begin{table}[t]
\centering
\caption{\textbf{Dataset Statistics of Agentic-MME.} textbf{Dataset Statistics of Agentic-MME.}  Specifically, \textbf{Total Images / Tools} denotes the unique visual inputs and tool invocations collected from human trajectories, while \textbf{PerDescri. Len} indicates the average token of the intent description per checkpoint.}
\vspace{-3mm}
\label{tab:agentic_mme_stats_premium}
\footnotesize
\aboverulesep=0pt 
\belowrulesep=0pt 
\renewcommand{\arraystretch}{1.35}
\begin{tabularx}{\textwidth}{>{\raggedright\arraybackslash}X >{\centering\arraybackslash}X | >{\raggedright\arraybackslash}X >{\centering\arraybackslash}X}
\toprule
\textbf{Task Difficulty} & \textbf{Share} & \textbf{Ckpts / Task} & \textbf{Tools / Task} \\
\midrule
Level 1 (Easy) & 48.6\% & 2.89 & 1.21 \\
Level 2 (Mid)  & 32.1\% & 4.64 & 2.42 \\
\rowcolor{rowhighlight} \textbf{Level 3 (Hard)} & \textbf{19.4\%} & \textbf{6.67} & \textbf{4.07} \\
\midrule
\midrule
\textbf{Key Property} & \textbf{Value} & \textbf{Key Property} & \textbf{Value} \\
\midrule
Total Images / Tools  & 430 / 899          & Small Cues ($<10\%$) & 226 (43.1\%) \\
Domains / Sub & 6 / 35             & Avg. Cue Area       & 35.8\% \\
Avg. Resolution       & $1952 \times 1747$ & Avg. $L_{prompt} / L_{ans}$ & 31.9 / 1.5 token\\
PerDescri. Len        & 16.38 tokens        & External Search & 29.4\% \\
\bottomrule
\end{tabularx}
\vspace{-4mm}
\end{table}
As illustrated in Figure~\ref{fig:data}(a), Agentic-MME features a highly diverse, hierarchical distribution across 6 major domains and 35 sub-categories with prompt diversity further highlighted by the word cloud in Figure~\ref{fig:data} (c). Table~\ref{tab:agentic_mme_stats_premium} summarizes the distribution of the  tasks across Levels 1--3 , alongside the average number of checkpoints and tool invocations in the human reference trajectories. The spatial footprint of visual evidence, showing that over 40\% of instances require recovering highly localized information (occupying $<10\%$ of the image area), which justifies the need for active tool manipulation over passive perception. Figure~\ref{fig:data}(b) reports the token length distribution of questions and answers. Crucially, the benchmark is grounded by over 2,000 fine-grained stepwise checkpoints. As shown in Figure~\ref{fig:data}(d) the checkpoint and tool calls density increases significantly from Level 1 to Level 3, consistent with the longer horizons in complex tasks. Notably, every checkpoint is accompanied by a detailed natural language description (averaging 16.38 tokens), ensuring precise interpretability of intent and failure modes for both human and automated evaluation.
\subsection{Unified Tool Interface and Execution Harness}
\label{sec:harness}
Agentic-MME evaluates agents from fully logged, replayable traces that include intermediate actions, returned observations, and generated artifacts (e.g., processed images and retrieved webpages). Our harness supports heterogeneous agents via two interaction modalities: atomic tool calls through a function interface, and sandboxed code execution for free-form visual manipulation. Making these interfaces comparable is non-trivial: models differ drastically in how they express image operations, how they name files, and how they handle I/O. Because smaller models often fail to override their native sandbox biases even when guided by rigorous system prompts. We enforce a strict artifact protocol that normalizes both modalities into a unified, auditable event stream. For code-writing agents, we execute programs in an instrumented sandbox that constrains and normalizes image I/O (e.g., save-path redirection), and use an AST-based tracer to extract canonical visual operations from executed code traces, enabling consistent checkpoint matching across coding styles. For retrieval, we provide API-based search (text, images, and page fetch) and store all queries and payloads in a cacheable format to reduce non-determinism and enable deterministic replay (Appendix~\ref{app:tools}).
\section{Experiments}
\label{sec:exp}
\vspace{-2mm}

\begin{table}[!t]
\centering
\caption{Comprehensive evaluation results on the Agentic-MME benchmark. We evaluate tool-augmented models under two distinct interaction interfaces: \textbf{Gen} (Code Generation mode, where models write sandboxed Python for visual transforms) and \textbf{Atm} (Atomic mode, interacting via structured function-calling APIs). Within each category, models are sorted in descending order based on their highest (Acc) across all modes. The top model in each category is highlighted with a small icon.}
\vspace{-3mm}
\label{tab:gaia_allmetrics_expanded_v3}
\small %
\setlength{\tabcolsep}{2.2pt} 
\renewcommand{\arraystretch}{1.1} 
\aboverulesep=0pt 
\belowrulesep=0pt 

\definecolor{colorNoTool}{HTML}{fbf3dd}   
\definecolor{colorOpen}{HTML}{ffeeb9}     
\definecolor{colorClosed}{HTML}{feeaab}   

\resizebox{0.98\textwidth}{!}{%
\begin{tabular}{@{} l @{\hspace{2pt}}c | ccc | cccc | cccc | cccc @{}}
\toprule
\multirow{2}{*}{\textbf{Model}} & \multirow{2}{*}{\textbf{Mode}}
& \multicolumn{3}{c|}{\textbf{Overall}}
& \multicolumn{4}{c|}{\textbf{Level 1}}
& \multicolumn{4}{c|}{\textbf{Level 2}}
& \multicolumn{4}{c}{\textbf{Level 3}} \\
\cmidrule(lr){3-5}\cmidrule(lr){6-9}\cmidrule(lr){10-13}\cmidrule(lr){14-17}
& & \textbf{Acc} & \textbf{S} & \textbf{V}
& \textbf{Acc} & \textbf{S} & \textbf{$V_{to}$} & \textbf{$V_{tr}$}
& \textbf{Acc} & \textbf{S} & \textbf{$V_{to}$} & \textbf{$V_{tr}$}
& \textbf{Acc} & \textbf{S} & \textbf{$V_{to}$} & \textbf{$V_{tr}$} \\
\midrule
\textbf{Human} & -- 
& \textbf{93.8} & -- & -- & \textbf{99.0} & -- & -- & -- & \textbf{92.6} & -- & -- & -- & \textbf{82.3} & -- & -- & -- \\

\midrule
\multicolumn{17}{l}{\cellcolor{colorNoTool}{\textit{No Tooluse Models }}} \\
\Xhline{0.5pt}
\makecell[l]{\textbf{Gemini 3}\\\textbf{pro-preview}} & --
& 30.2 & -- & -- & 42.9 & -- & -- & -- & 24.6 & -- & -- & -- & 7.5 & -- & -- & -- \\

\midrule
\multicolumn{17}{l}{\cellcolor{colorOpen}{\textit{Open-Source Models}}} \\
\midrule

\multirow{2}{*}{\makecell[l]{\textbf{Qwen 3}\\\textbf{Vl-235B}}\includegraphics[width=0.3cm]{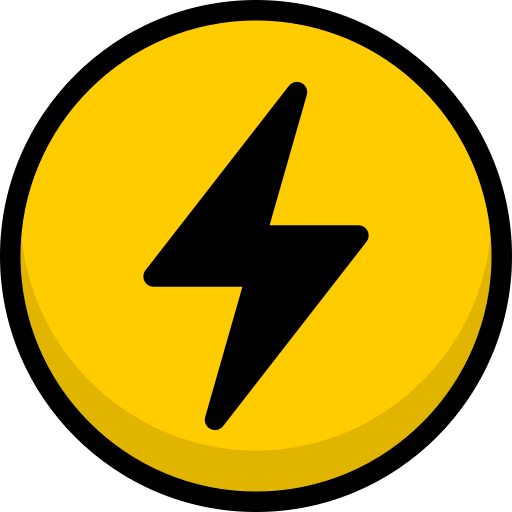}} & Gen
& 28.5 & 20.8 & 17.3 & 43.9 & 28.6 & 17.8 & 18.2 & 23.3 & 33.0 & 20.0 & 19.3 &  2.5 & 10.1 & 12.1 & 15.2 \\
& Atm
& \textbf{34.9} & 20.5 & 35.7 & 50.0 & 57.1 & 27.7 & 30.2 & 26.9 & 30.1 & 40.7 & 40.5 & 10.1 &  8.1 & 48.8 & 48.4 \\
\addlinespace[0.15em]

\multirow{2}{*}{\makecell[l]{\textbf{Qwen3 Vl}\\\textbf{32B-think}}} & Gen
& 27.4 &  8.7 & 36.5 & 42.2 & 21.4 & 66.4 &  9.7 & 19.5 & 15.7 & 37.9 & 18.7 &  4.6 &  4.8 & 30.4 & 12.6 \\
& Atm
& 29.9 &  9.8 & 12.7 & 46.3 & 21.4 & 17.1 &  9.4 & 22.5 & 17.0 & 15.5 & 17.5 &  4.6 &  6.1 & 14.3 & 14.2 \\
\addlinespace[0.15em]

\multirow{2}{*}{\makecell[l]{\textbf{Qwen3 Vl}\\\textbf{8B-think}}} & Gen
& 26.3 & 10.3 & 22.2 & 40.4 & 21.4 & 29.5 &  8.7 & 17.3 & 16.4 & 31.4 & 18.2 &  9.1 &  6.7 & 19.6 & 13.9 \\
& Atm
& 26.6 & 10.2 & 12.3 & 45.6 & 25.0 & 10.6 &  9.2 & 16.5 & 20.8 & 13.4 & 18.8 &  3.3 &  5.9 &  8.1 & 13.2 \\
\addlinespace[0.15em]

\multirow{2}{*}{\textbf{Deepeyesv2}} & Gen
& 22.5 &  4.5 & 12.9 & 34.3 &  0.0 & 24.5 &  6.6 & 13.5 &  7.0 & 17.2 &  0.6 &  7.5 &  2.9 &  7.0 &  0.8 \\
& Atm
& 25.2 &  5.1 & 26.5 & 38.9 &  0.0 & 24.4 & 14.4 & 17.0 &  9.3 & 44.8 & 11.5 &  3.8 &  2.0 & 42.1 &  9.4 \\
\addlinespace[0.15em]

\multirow{2}{*}{\textbf{Thyme-rl}} & Gen
& 18.0 &  0.4 & 36.9 & 27.7 &  0.0 & 63.3 & 13.0 & 10.5 &  1.0 & 38.8 &  4.2 &  6.2 &  0.0 & 35.6 &  4.7 \\
& Atm
& 23.0 &  1.8 & 25.1 & 32.4 &  0.0 & 23.8 & 13.7 & 20.7 &  2.9 & 44.0 & 13.3 &  2.5 &  1.1 & 54.7 & 10.9 \\

\midrule
\multicolumn{17}{l}{\cellcolor{colorClosed}{\textit{Closed-Source Models}}} \\
\midrule

\multirow{2}{*}{\makecell[l]{\textbf{Gemini 3}\\\textbf{pro-preview}}\includegraphics[width=0.3cm]{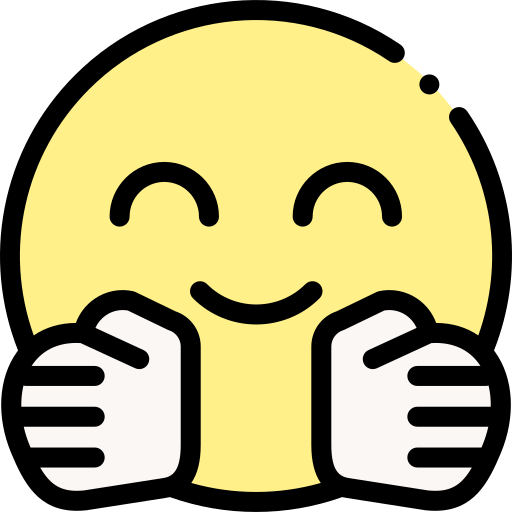}} & Gen
& 49.5 & 29.7 & 45.8 & 64.8 & 42.9 & 53.0 & 38.1 & 41.8 & 41.8 & 48.6 & 29.5 & 23.0 & 26.1 & 54.1 & 23.0 \\
& Atm
& \textbf{56.3} & 29.0 & 64.1 & 70.6 & 50.0 & 77.9 & 43.2 & 47.4 & 37.8 & 75.9 & 36.1 & 33.3 & 18.2 & 87.5 & 28.8 \\
\addlinespace[0.15em]

\multirow{2}{*}{\textbf{Gemini 3-flash}} & Gen
& 47.7 & 27.3 & 51.0 & 62.8 & 50.0 & 72.8 & 28.1 & 40.0 & 39.4 & 62.1 & 30.6 & 21.6 & 26.1 & 63.1 & 20.9 \\
& Atm
& 52.2 & 26.4 & 59.4 & 67.6 & 57.1 & 67.0 & 35.2 & 43.8 & 46.3 & 70.8 & 31.5 & 27.2 & 18.8 & 67.5 & 24.8 \\
\addlinespace[0.15em]

\multirow{2}{*}{\textbf{Qwen3.5-plus}} & Gen
& 44.5 & 27.3 & 45.0 & 62.1 & 42.9 & 69.6 & 35.8 & 39.8 & 40.3 & 56.1 & 31.5 & 18.3 & 25.5 & 35.9 & 19.3 \\
& Atm
& 47.1 & 29.0 & 58.8 & 61.8 & 35.7 & 71.9 & 37.4 & 39.6 & 41.7 & 70.7 & 37.2 & 22.8 & 17.5 & 75.5 & 26.2 \\
\addlinespace[0.15em]

\multirow{2}{*}{\textbf{Kimi-k2.5}} & Gen
& 35.0 & 28.0 & 16.8 & 50.8 & 64.3 & 21.7 & 27.3 & 29.0 & 38.2 &  9.1 & 16.9 &  8.8 & 22.5 &  5.1 & 13.4 \\
& Atm
& 44.0 & 26.2 & 47.3 & 61.7 & 42.9 & 60.9 & 34.5 & 39.0 & 40.3 & 51.9 & 27.7 & 20.9 & 26.2 & 60.4 & 17.6 \\
\addlinespace[0.15em]

\multirow{2}{*}{\textbf{GPT 5-mini}} & Gen
& 33.5 & 22.3 & 56.1 & 47.8 & 42.9 & 82.7 & 35.3 & 29.2 & 29.4 & 70.8 & 27.3 &  8.2 & 18.7 & 72.0 & 24.2 \\
& Atm
& 37.7 & 20.8 & 53.1 & 50.5 & 35.7 & 58.3 & 38.1 & 31.9 & 24.7 & 78.4 & 30.7 & 14.3 & 15.8 & 79.4 & 28.9 \\
\addlinespace[0.15em]

\multirow{2}{*}{\textbf{GPT 5.2}} & Gen
& 31.2 & 20.3 & 10.4 & 45.3 & 21.4 &  7.6 & 21.1 & 24.6 & 33.0 &  6.7 & 15.8 & 13.5 & 15.4 &  6.9 & 11.6 \\
& Atm
& 35.0 & 17.4 & 58.1 & 49.0 & 35.7 & 79.0 & 39.1 & 26.1 & 22.6 & 71.0 & 24.2 & 16.7 & 11.4 & 72.6 & 12.5 \\
\bottomrule
\end{tabular}
}
\vspace{-8pt}
\end{table}
\subsection{Experimental Setup}
\noindent\textbf{Models and baselines.} We evaluate a diverse set of open-source and close-source multimodal models on Agentic-MME. Human reference baseline is obtained by averaging the scores of three independent human solvers, who are allowed to use both search engines and pure perception models as auxiliary tools. We also report No-tool baselines where models answer from the raw images without invoking any tools.
For tool-augmented agents, we evaluate representative open-source models Thyme-rl\citep{zhang2025thyme}, Deepeyesv2\citep{hong2025deepeyesv2}, Qwen3-VL-235B, Qwen3-VL-8B-thinking, Qwen~3~VL-32B-thinking\citep{bai2025qwen3} and closed-source models like Gemini3 famil\citep{team2023gemini}, Kimi-k2.5\citep{team2026kimi}, GPT-5.2\citep{openai2025thinking}, Qwen3.5-plus\citep{qwen35blog} as reported in \cref{tab:gaia_allmetrics_expanded_v3}.

\vspace{1mm}
\noindent\textbf{Two tool interfaces: Code and Atomic.}
A central design goal of Agentic-MME is to benchmark agentic capability across heterogeneous implementations.
We therefore evaluate each tool-using model under two interfaces supported by our unified harness (Sec.~\ref{sec:harness}):
Code mode (\textbf{Gen}), where the model writes sandboxed Python to perform visual transforms; and
Atomic mode (\textbf{Atm}), where the model interacts via structured function calls following the schemas (Appx.~\ref{app:tools}). This controlled comparison directly tests whether tool competence generalizes across interfaces, rather than being tied to one training format.

\noindent\textbf{Evaluation protocol.} All evaluations are conducted on fully logged, replayable traces. Inspired by CodeV~\citep{hou2025codev}, we adopt an MLLM-as-a-Judge approach to verify the intermediate visual artifacts ($V_{true}$). Similarly, search checkpoints are assessed by an LLM judge that evaluates the agent's retrieved URLs, keywords, and intermediate thinking processes against human-annotated targets. We use gpt-4o-mini\citep{hurst2024gpt} as the primary judge model in our experiments.

\subsection{Main Results on Agentic-MME}
\label{sec:exp_main}
\noindent\textbf{Finding 1: All models fall far below human performance, with
a sharp accuracy drop on Level-3.} Human solvers reach 93.8\% overall and remain strong even on the hardest split (L3: 82.3\%). In contrast, every model shows a significant decline as task complexity grows. The best-performing system, Gemini~3~Pro (Atm), achieves 56.3\% overall but only 33.3\% on Level-3. For comparison, Gemini~3~Pro \emph{without any tools} scores 42.9\% on Level-1 through passive perception alone, but drops to 7.5\% on Level-3; with full tool access, L3 accuracy rises to 33.3\%, a 4.4$\times$ improvement that demonstrates tools are essential for hard tasks. Nevertheless, the large remaining gap to human performance (33.3\% vs.\ 82.3\% on L3) shows that current agents still lack the
multi-step planning and reliable tool execution that complex real-world workflows demand.

\vspace{1mm}
\noindent\textbf{Finding 2: Open-source models lag behind closed-source
models, primarily in search and planning.}
Closed-source models consistently outperform open-source alternatives
across all levels. This gap is most visible on Level-3, where Qwen3~VL-235B drops to
10.1\% and Thyme-rl collapses to 2.5\%. The S-axis reveals the mechanism behind this gap: Thyme-rl and Deepeyesv2 achieve S scores below 5\%, indicating near-total failure
in formulating search queries and extracting useful intermediate
answers. Qwen3~VL-235B reaches S$\approx$20\%, still well below closed-source
models. In short, current open-source models can learn to \emph{invoke} tools
but have not yet acquired the retrieval planning needed for reliable
multi-step problem solving.

\vspace{1mm}
\noindent\textbf{Finding 3: Structured tool APIs outperform code generation,
but code mode has untapped potential.}Across all models, Atm mode matches or exceeds Gen mode. Process metrics help explain this gap. GPT~5.2 in Gen mode seldomly writes image-processing code ($V_{\mathit{to}} \leq 7.6\%$ across all levels), resulting in an
overall V of only 10.4; switching to Atm raises $V_{\mathit{to}}$ above 70\% and V to 58.1.
Kimi-k2.5 shows a similar pattern. A plausible explanation is that code generation imposes a higher cognitive burden: models must handle library imports, manage file
I/O, deal with error recovery, and compose multi-step programs correctly. However, the code mode offers unique advantages that structured APIs cannot easily replicate: arbitrary composition of operations, custom transformations beyond predefined tools. In our experiments, models like Gemini~3~Flash (Gen) already achieve competitive results (47.7\% overall).How to bridge the reliability gap while preserving the flexibility
of code generation remains a valuable open question that needs further investigation.

\vspace{1mm}
\noindent\textbf{Finding 4: Models call tools eagerly but often produce
wrong results.} Some agents invoke tools aggressively but generate incorrect
outputs: Thyme-rl (Gen, L1) reaches $V_{\mathit{tool}}=63.3$ while
$V_{\mathit{true}}$ is only 13.0; Qwen3~VL-32B-think (Gen, L1) shows
$V_{\mathit{tool}}=66.4$ vs.\ $V_{\mathit{true}}=9.7$.
These models frequently crop or transform the wrong region, wasting
interaction budget and injecting misleading artifacts.
In contrast, Qwen3~VL-235B (Atm, L3) shows
$V_{\mathit{tool}}=48.8$ and $V_{\mathit{true}}=48.4$, nearly
identical, indicating that its tool parameterization is reliable. Agentic-MME's process-level checkpoints make it possible to pinpoint where and why model fails.
\subsection{Further Analysis}
\label{sec:fur}
A meaningful benchmark should genuinely require the capabilities it
claims to test We conduct three validation experiments to confirm that Agentic-MME
(i) cannot be solved with text only, (ii) truly demands
active tool use rather than passive perception, and (iii) provides
high-quality stepwise annotations that capture actionable problem-solving structure. 

\vspace{1mm}
\noindent\textbf{Visual grounding is essential}
Removing all images yields near-zero accuracy (Gemini~3~Flash: 2.63\%;
GPT-5-mini: 1.44\%), confirming that Agentic-MME is free from data leakage and that each task genuinely requires visual evidence.

\vspace{1mm}
\noindent\textbf{Tasks genuinely require active tool use.}
To validate that Agentic-MME rewards agentic behavior rather than
strong passive perception, we compare four settings:
Perception-only, Image-only, Search-only, and Full
(\cref{tab:tool_ablation}).
For both models, the full system consistently outperforms all restricted settings, confirming that the benchmark is designed to require both Visual and Knowledge
Expansion. The Level-3 split offers further validation that our hardest tasks
capture meaningful capability synergy.
For Qwen3~VL-235B, adding image tools alone actually \emph{hurts}
Level-3 (7.41\%$\to$6.25\%), while search-only helps modestly
(7.41\%$\to$11.11\%). However combining both yields 19.23\%, gain that far
exceeds the sum of individual effects. This super-additive pattern validates our Level-3 design: these tasks are hard to solve by either capability in isolation. Gemini~3~Flash shows the same trend on Level-3 tasks. The results also suggest that unreliable visual tools can be counterproductive without search-based verification.
\begin{table}[t]
\centering
\caption{\textbf{Tool availability analysis.} We reprot Acc in different level tasks}
\vspace{-3mm}
\label{tab:tool_ablation}
\small
\setlength{\tabcolsep}{6.9pt}
\renewcommand{\arraystretch}{1.45}
\setlength{\aboverulesep}{0pt}
\setlength{\belowrulesep}{0pt}

\begin{tabular}{l cccc cccc} 
\toprule
\rowcolor{CadetBlue!20}
& \multicolumn{4}{c}{\textbf{Gemini~3~Flash}} & \multicolumn{4}{c}{\textbf{Qwen~3~VL-235B}} \\
\cmidrule(lr){2-5}\cmidrule(lr){6-9}
\rowcolor{CadetBlue!20}
\multirow{-2}{*}{\textbf{Setting}} & \textbf{Overall} & \textbf{L1} & \textbf{L2} & \textbf{L3} & \textbf{Overall} & \textbf{L1} & \textbf{L2} & \textbf{L3} \\
\midrule
Perception-only  & 39.82 & 55.84 & 30.10 & 15.24 & 31.94 & 48.44 & 23.13 &  7.41 \\
Image-only       & 47.61 & 64.58 & 38.90 & 18.79 & 36.93 & 57.67 & 25.58 &  6.25 \\
Search-only      & 46.68 & 65.62 & 34.33 & 22.22 & 37.10 & 52.08 & 31.34 & 11.11 \\
\rowcolor{customblue!20} 
\textbf{Full (Img+Search)}& \textbf{52.24} & \textbf{67.60} & \textbf{43.77} & \textbf{27.19} & \textbf{42.85} & \textbf{58.82} & \textbf{34.27} & \textbf{19.23} \\
\bottomrule
\end{tabular}
\vspace{2mm}
\end{table}

\vspace{1mm}
\noindent\textbf{Stepwise annotations are high-quality and effective}
A core contribution of Agentic-MME is the human-annotated reference
trajectories with over 2,000 stepwise checkpoints.
We validate their quality by feeding them to models as guidance and measuring whether performance improves(\cref{tab:oracle_guidance}).
\textit{+Visual Cues} supplies the ground-truth intermediate visual
artifacts from our V-axis annotations (e.g., the correctly cropped
region); \textit{+Stepwise Guidance} further provides the per checkpoint description derived from our annotation. To prevent leakage, we mask all answer-related keywords in the guidance.
Both forms of annotation produce consistent improvements.
Stepwise guidance yields substantially improvements, validating that our human trajectories capture actionable planning structure. Notably, the stronger model benefits more from the annotations. A plausible explanation is that the stronger model has sufficient planning ability to incorporate the extra visual evidence into its reasoning. Crucially, performance does not saturate to near-perfect scores, especially on Level-3, despite our step-wise verification during data collection. This reflects the gap between step-by-step verification and autonomous execution. Even given a perfect blueprint, agents must still write API calls, track long contexts, and avoid compounding errors—proving that continuous execution is far harder than isolated perception.
\begin{table}[t]
\centering
\caption{\textbf{Oracle guidance study.} We report acc in different level tasks.}
\vspace{-3mm}
\label{tab:oracle_guidance}
\small
\setlength{\tabcolsep}{6.5pt}
\renewcommand{\arraystretch}{1.45}

\setlength{\aboverulesep}{0pt}
\setlength{\belowrulesep}{0pt}

\begin{tabular}{l cccc cccc} 
\toprule
\rowcolor{CadetBlue!20}
& \multicolumn{4}{c}{\textbf{Gemini~3~Flash}} & \multicolumn{4}{c}{\textbf{Qwen~3~VL-235B}} \\
\cmidrule(lr){2-5}\cmidrule(lr){6-9}
\rowcolor{CadetBlue!20}
\multirow{-2}{*}{\textbf{Setting}} & \textbf{Overall} & \textbf{L1} & \textbf{L2} & \textbf{L3} & \textbf{Overall} & \textbf{L1} & \textbf{L2} & \textbf{L3} \\
\midrule
Full (baseline)
  & 52.24 & 67.60 & 43.77 & 27.19
  & 42.85 & 58.82 & 34.27 & 19.23 \\
+ Visual Cues
  & 58.37 & 76.56 & 45.52 & 36.25
  & 49.38 & 72.11 & 38.17 & 13.75 \\
\rowcolor{customblue!20} 
\textbf{+ Stepwise Guidance}
  & \textbf{76.21} & \textbf{89.58} & \textbf{71.94} & \textbf{51.25}
  & \textbf{72.80} & \textbf{88.59} & \textbf{66.67} & \textbf{46.91} \\
\bottomrule
\end{tabular}
\vspace{-4mm}
\end{table}
\definecolor{colorOpen}{HTML}{ffeeb9}
\definecolor{colorClosed}{HTML}{feeaab}

\begin{figure}[t]
\begin{minipage}[t]{0.54\textwidth}
    \centering
    \vspace{0pt}
    \begin{minipage}[t]{0.92\linewidth}
        \centering
        \footnotesize
        \captionof{table}{Efficiency analysis. Avg. calls per task \\ and Overthink (OT). We report human ref.: \\ 2.15 calls per task.}

        \setlength{\tabcolsep}{9pt}
        \setlength{\aboverulesep}{0pt}
        \setlength{\belowrulesep}{0pt}
        {\renewcommand{\arraystretch}{1.10}
        \begin{tabular}{l cc cc}
            \toprule
            \rowcolor{CadetBlue!20}
            & \multicolumn{2}{c}{\textbf{Gen}}
            & \multicolumn{2}{c}{\textbf{Atm}} \\
            \cmidrule(lr){2-3}\cmidrule(lr){4-5}
            \rowcolor{CadetBlue!20}
            \multirow{-2}{*}{\textbf{Model}}
            & \textbf{Calls} & \textbf{OT}
            & \textbf{Calls} & \textbf{OT} \\
            \midrule
            \multicolumn{5}{l}{\cellcolor{colorOpen}\textit{Open-Source}} \\
            DeepeyesV2    & 1.98  & 0.00 & 1.95 & 0.00 \\
            Qwen3-235B    & 8.66  & 4.03 & 3.31 & 1.54 \\
            \midrule
            \multicolumn{5}{l}{\cellcolor{colorClosed}\textit{Closed-Source}} \\
            Gemini~3~Pro  & 9.28  & 2.26 & 4.66 & 0.80 \\
            Kimi-k2.5     & 9.43  & 2.31 & 4.48 & 0.74 \\
            GPT~5.2       & 5.50  & 1.06 & 2.98 & 0.26 \\
            GPT-5-mini    & \textbf{12.13} & 5.64 & \textbf{7.22} & 3.36 \\
            Qwen3.5-plus  & 7.72  & 3.59 & 5.24 & 2.44 \\
            \bottomrule
        \end{tabular}
        }
        \label{tab:overthink}
    \end{minipage}
\end{minipage}
\hspace{1mm}
\begin{minipage}[t]{0.43\textwidth}
    \centering
    \vspace{0pt}
    \begin{minipage}[t]{0.92\linewidth}
        \centering
        \footnotesize
        \captionof{table}{Evaluation robustness on \\ Gemini~3~Pro (Atm). We report \\ overall scores.}

        \setlength{\tabcolsep}{6pt}
        \setlength{\aboverulesep}{0pt}
        \setlength{\belowrulesep}{0pt}
        {\renewcommand{\arraystretch}{2.43}
        \begin{tabular}{l ccc}
            \toprule
            \rowcolor{CadetBlue!20}
            \textbf{Judge}
            & \textbf{Acc} & \textbf{S} & \textbf{V} \\
            \midrule
            GPT-5-mini       & 56.28 & 28.44 & 63.27 \\
            Gem-2.5-Flash    & 56.28 & 29.56 & 65.64 \\
            GPT-4o-mini      & 56.28 & 29.00 & 64.08 \\
            \midrule
            \rowcolor{customblue!20}
            Human Expert     & 56.28 & 28.67 & 64.84 \\
            \bottomrule
        \end{tabular}
        }

        \label{tab:judge_robust}
    \end{minipage}
\end{minipage}
\vspace{2mm}
\end{figure}

\vspace{1mm}
\noindent\textbf{Models systematically overthink.} Beyond accuracy, we quantify how efficiently models solve tasks relative to human expert trajectories. \cref{tab:overthink} reports per-task average tool calls and the
Overthink ratio. Two patterns emerge. First, Atm mode is universally more efficient than Gen mode. Second, neither interaction extreme guarantees high accuracy. DeepeyesV2 under-utilizes tools (OT=0) yielding low accuracy (22.5\%), while GPT-5-mini severely over-explores (12.13 calls/task) but only reaches 33.5\% due to redundant trial-and-error. Gemini~3~Pro achieves the highest accuracy (56.3\%) by striking a balance, demonstrating that \emph{focused, reliable} tool execution matters far more than exhaustive exploration.

\vspace{1mm}
\noindent\textbf{Evaluation robustness and human consistency} We verify consistency by re-evaluating Gemini~3~Pro (Atm) with three judges and human experts (\cref{tab:judge_robust}). S/V scores vary by a very small difference across judges, and human expert scores fall within the range, confirming that our structured checkpoint design aligns automated evaluation with human judgment.

\subsection{Fine-Grained Error Analysis}
\begin{figure}[t]
  \centering
  \includegraphics[width=0.9\linewidth, trim=5 5 5 5, clip]{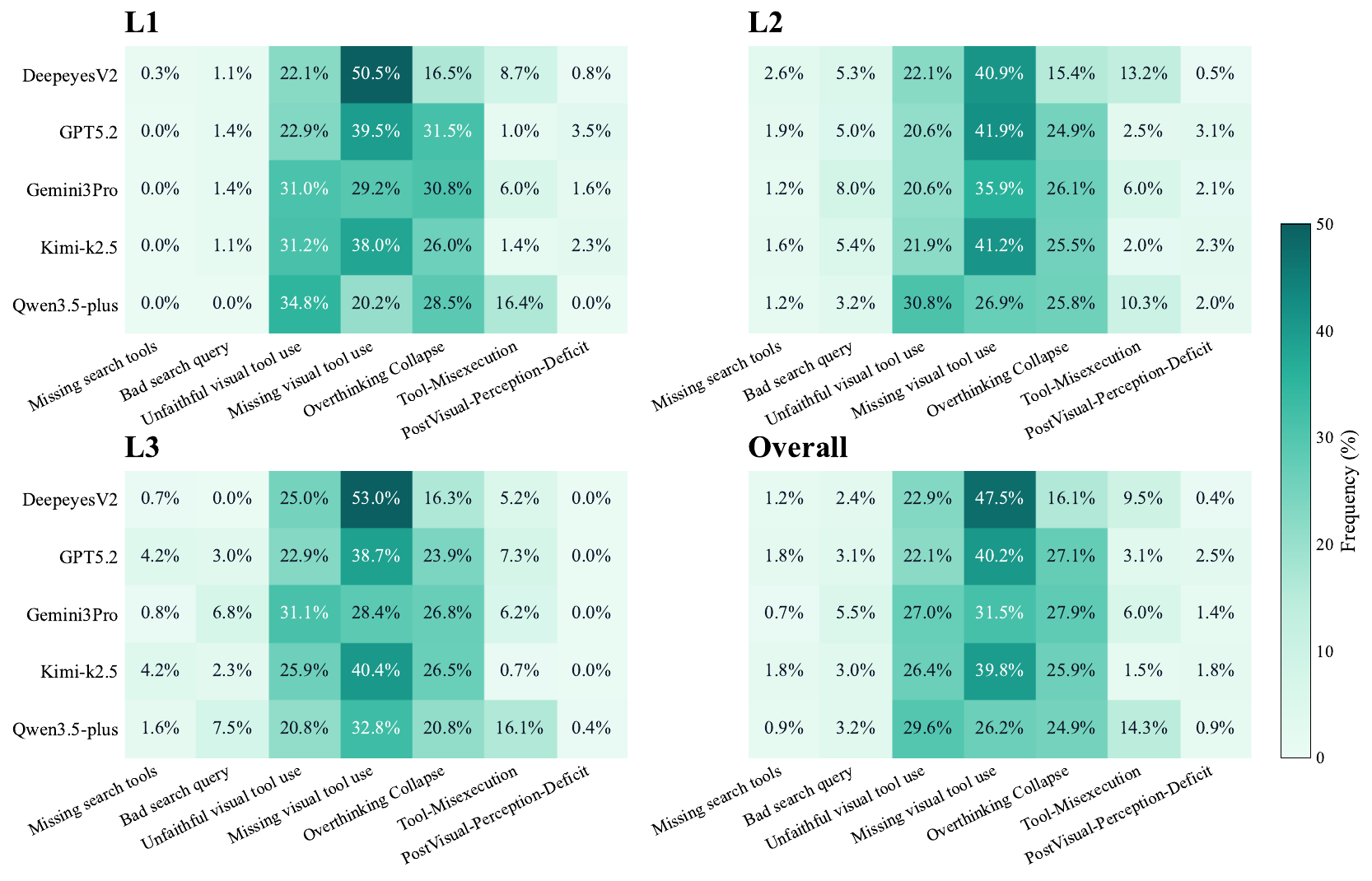}
  \caption{Fine-Grained Error Analysis. The heatmap illustrates the frequency of seven failure modes across different difficulty levels, averaged over both Code (Gen) and Atomic (Atm) execution modes. Darker colors denote higher frequencies.}
  \label{fig:error_analysis}
  \vspace{-4mm}
\end{figure}
To deeply diagnose bottlenecks, we categorize failures into seven modes (\cref{fig:error_analysis}). Four distinct patterns emerge: First, \textbf{reluctance to act} dominates; models frequently default to passive guessing rather than actively manipulating images, accounting for up to $\sim$50\% of errors. Second, strong agents are highly prone to \textbf{overthinking collapse}, trapping themselves in redundant tool-invocation loops. Third, \textbf{unfaithful execution} remains persistent, as models often crop irrelevant regions and inject noisy artifacts. Finally, while free-form code generation suffers from frequent syntax errors (\textit{Tool-Misexecution}), structured Atomic APIs effectively eliminate these low-level derailments.

\section{Related Work}
\vspace{-2mm}
\noindent\textbf{Tool-augmented visual reasoning.} While traditional evaluations target static multimodal inputs~\citep{fu2023mme,yue2024mmmu,li2023seed,mmevideoocr,yu2023mm}, recent benchmarks explore active multi-tool execution~\citep{wang2024gta,ma2024m} and visual manipulation~\citep{guo2025beyond,li2025tir,realunify}. However, these settings treat open-web retrieval as peripheral (e.g., \texttt{google\_search} constitutes $<7\%$ of calls for o3/GPT-5), failing to thoroughly assess the synergy between \textit{Visual} and \textit{Knowledge Expansion} in realistic workflows (Table~\ref{tab:compare}).

\vspace{1mm}
\noindent\textbf{Multimodal search and process-aware evaluation.} omplementary lines of work focus on open-world information seeking, evaluating how models incorporate vision in the loop of search~\citep{jiang2024mmsearch,tao2025mmsearch} and multimodal web browsing~\citep{li2025mm}. However, as highlighted by CodeV~\citep{hou2025codev}, relying solely on the correctness of final answers can easily mask unfaithful tool execution. Consequently, while recent benchmarks advance textual provenance tracking, intermediate visual artifacts often remain unverified, despite a growing consensus on the need for strict, step-wise process verification in agent evaluations~\citep{froger2026gaia2,zhang2026adaptmmbench}. Similarly, recent multimodal deep-research frameworks have made inspiring strides in long-form report synthesis~\citep{huang2026vision,huang2026mmdeepresearch}. However, because their primary objective is knowledge retrieval, their visual operations are largely restricted to pre-processing (e.g., parsing webpage screenshots), lacking a comprehensive suite of active visual manipulation tools.

\section{Conclusion}
\vspace{-2mm}
We introduce Agentic-MME, a process-verified benchmark designed to systematically evaluate the deep synergy between active visual manipulation (Visual Expansion) and open-web retrieval (Knowledge Expansion) in multimodal agents. Moving beyond the opaque final-answer grading prevalent in existing evaluations, we contribute a unified execution harness that supports heterogeneous tool interfaces, grounded in over 2,000 human-annotated stepwise checkpoints. This dual-axis framework enables granular auditing of intermediate tool intent, visual artifact faithfulness, and execution efficiency. 

Our extensive evaluation exposes a critical gap between frontier models and human performance, particularly in complex workflows. We demonstrate that while current models can execute simple sequential tool-chaining, they severely struggle with advanced synergistic tasks—such as resolving visual ambiguity through fuzzy search and conducting iterative hypothesis verification across modalities. By pinpointing these precise bottlenecks, including unfaithful tool execution and redundant "overthinking" loops, Agentic-MME transcends standard capability testing. It provides a rigorous, diagnostic roadmap for developing the next generation of robust, long-horizon multimodal agents capable of intertwined visual and knowledge reasoning.
\clearpage  


%
%
\bibliographystyle{unsrtnat}
\bibliography{main}

@article{team2023gemini,
  title={Gemini: a family of highly capable multimodal models},
  author={Team, Gemini and Anil, Rohan and Borgeaud, Sebastian and Wu, Yonghui and Alayrac, Jean-Baptiste and Yu, Jiahui and Soricut, Radu and Schalkwyk, Johan and Dai, Andrew M and Hauth, Anja and others},
  journal={arXiv preprint arXiv:2312.11805},
  year={2023}
}

@article{li2023seed,
  title={Seed-bench: Benchmarking multimodal llms with generative comprehension},
  author={Li, Bohao and Wang, Rui and Wang, Guangzhi and Ge, Yuying and Ge, Yixiao and Shan, Ying},
  journal={arXiv preprint arXiv:2307.16125},
  year={2023}
}

@article{fu2023mme,
  title={MME: A Comprehensive Evaluation Benchmark for Multimodal Large Language Models},
  author={Fu, Chaoyou and Chen, Peixian and Shen, Yunhang and Qin, Yulei and Zhang, Mengdan and Lin, Xu and Qiu, Zhenyu and Lin, Wei and Yang, Jinrui and Zheng, Xiawu and others},
  journal={arXiv:2306.13394},
  year={2023}
}

@article{zheng2025deepeyes,
  title={DeepEyes: Incentivizing" Thinking with Images" via Reinforcement Learning},
  author={Zheng, Ziwei and Yang, Michael and Hong, Jack and Zhao, Chenxiao and Xu, Guohai and Yang, Le and Shen, Chao and Yu, Xing},
  journal={arXiv preprint arXiv:2505.14362},
  year={2025}
}

@misc{openai2025thinking,
  author = {{OpenAI}},
  title = {Thinking with images},
  year = {2025},
  howpublished = {\url{https://openai.com/index/thinking-with-images/}},
}

@article{hong2025deepeyesv2,
  title={DeepEyesV2: Toward Agentic Multimodal Model},
  author={Hong, Jack and Zhao, Chenxiao and Zhu, ChengLin and Lu, Weiheng and Xu, Guohai and Yu, Xing},
  journal={arXiv preprint arXiv:2511.05271},
  year={2025}
}

@article{zhang2025skywork,
  title={Skywork-R1V4: Toward Agentic Multimodal Intelligence through Interleaved Thinking with Images and DeepResearch},
  author={Zhang, Yifan and Hu, Liang and Sun, Haofeng and Wang, Peiyu and Wei, Yichen and Yin, Shukang and Pei, Jiangbo and Shen, Wei and Xia, Peng and Peng, Yi and others},
  journal={arXiv preprint arXiv:2512.02395},
  year={2025}
}

@article{wang2024gta,
  title={GTA: a benchmark for general tool agents},
  author={Wang, Jize and Zerun, Ma and Li, Yining and Zhang, Songyang and Chen, Cailian and Chen, Kai and Le, Xinyi},
  journal={Advances in Neural Information Processing Systems},
  volume={37},
  pages={75749--75790},
  year={2024}
}

@inproceedings{ma2024m,
  title={m \& m’s: A benchmark to evaluate tool-use for m ulti-step m ulti-modal tasks},
  author={Ma, Zixian and Huang, Weikai and Zhang, Jieyu and Gupta, Tanmay and Krishna, Ranjay},
  booktitle={European Conference on Computer Vision},
  pages={18--34},
  year={2024},
  organization={Springer}
}

@article{li2025tir,
  title={TIR-Bench: A Comprehensive Benchmark for Agentic Thinking-with-Images Reasoning},
  author={Li, Ming and Zhong, Jike and Zhao, Shitian and Zhang, Haoquan and Lin, Shaoheng and Lai, Yuxiang and Wei, Chen and Psounis, Konstantinos and Zhang, Kaipeng},
  journal={arXiv preprint arXiv:2511.01833},
  year={2025}
}

@article{guo2025beyond,
  title={Beyond seeing: Evaluating multimodal llms on tool-enabled image perception, transformation, and reasoning},
  author={Guo, Xingang and Tyagi, Utkarsh and Gosai, Advait and Vergara, Paula and Park, Jayeon and Montoya, Ernesto Gabriel Hern{\'a}ndez and Zhang, Chen Bo Calvin and Hu, Bin and He, Yunzhong and Liu, Bing and others},
  journal={arXiv preprint arXiv:2510.12712},
  year={2025}
}

@article{su2025thinking,
  title={Thinking with images for multimodal reasoning: Foundations, methods, and future frontiers},
  author={Su, Zhaochen and Xia, Peng and Guo, Hangyu and Liu, Zhenhua and Ma, Yan and Qu, Xiaoye and Liu, Jiaqi and Li, Yanshu and Zeng, Kaide and Yang, Zhengyuan and others},
  journal={arXiv preprint arXiv:2506.23918},
  year={2025}
}

@article{zhang2025thyme,
  title={Thyme: Think beyond images},
  author={Zhang, Yi-Fan and Lu, Xingyu and Yin, Shukang and Fu, Chaoyou and Chen, Wei and Hu, Xiao and Wen, Bin and Jiang, Kaiyu and Liu, Changyi and Zhang, Tianke and others},
  journal={arXiv preprint arXiv:2508.11630},
  year={2025}
}

@article{hou2025codev,
  title={CodeV: Code with Images for Faithful Visual Reasoning via Tool-Aware Policy Optimization},
  author={Hou, Xinhai and Xu, Shaoyuan and Biyani, Manan and Li, Moyan and Liu, Jia and Hollon, Todd C and Wang, Bryan},
  journal={arXiv preprint arXiv:2511.19661},
  year={2025}
}

@article{li2025mm,
  title={Mm-browsecomp: A comprehensive benchmark for multimodal browsing agents},
  author={Li, Shilong and Bu, Xingyuan and Wang, Wenjie and Liu, Jiaheng and Dong, Jun and He, Haoyang and Lu, Hao and Zhang, Haozhe and Jing, Chenchen and Li, Zhen and others},
  journal={arXiv preprint arXiv:2508.13186},
  year={2025}
}

@article{tao2025mmsearch,
  title={Mmsearch-plus: Benchmarking provenance-aware search for multimodal browsing agents},
  author={Tao, Xijia and Teng, Yihua and Su, Xinxing and Fu, Xinyu and Wu, Jihao and Tao, Chaofan and Liu, Ziru and Bai, Haoli and Liu, Rui and Kong, Lingpeng},
  journal={arXiv preprint arXiv:2508.21475},
  year={2025}
}

@article{chen2025mindwatcher,
  title={MindWatcher: Toward Smarter Multimodal Tool-Integrated Reasoning},
  author={Chen, Jiawei and Shen, Xintian and Zheng, Lihao and Shao, Zhenwei and Zhang, Hongyuan and Yu, Pengfei and Rao, Xudong and Mao, Ning and Liu, Xiaobo and Wen, Lian and others},
  journal={arXiv preprint arXiv:2512.23412},
  year={2025}
}

@article{jiang2024mmsearch,
  title={Mmsearch: Benchmarking the potential of large models as multi-modal search engines},
  author={Jiang, Dongzhi and Zhang, Renrui and Guo, Ziyu and Wu, Yanmin and Lei, Jiayi and Qiu, Pengshuo and Lu, Pan and Chen, Zehui and Fu, Chaoyou and Song, Guanglu and others},
  journal={arXiv preprint arXiv:2409.12959},
  year={2024}
}

@article{zhang2026adaptmmbench,
  title={AdaptMMBench: Benchmarking Adaptive Multimodal Reasoning for Mode Selection and Reasoning Process},
  author={Zhang, Xintong and Zhang, Xiaowen and Wu, Jongrong and Gao, Zhi and Yan, Shilin and Diao, Zhenxin and Gao, Kunpeng and Chen, Xuanyan and Wu, Yuwei and Jia, Yunde and others},
  journal={arXiv preprint arXiv:2602.02676},
  year={2026}
}

@article{wei2026zooming,
  title={Zooming without Zooming: Region-to-Image Distillation for Fine-Grained Multimodal Perception},
  author={Wei, Lai and He, Liangbo and Lan, Jun and Dong, Lingzhong and Cai, Yutong and Li, Siyuan and Zhu, Huijia and Wang, Weiqiang and Kong, Linghe and Wang, Yue and others},
  journal={arXiv preprint arXiv:2602.11858},
  year={2026}
}

@article{froger2026gaia2,
  title={Gaia2: Benchmarking LLM Agents on Dynamic and Asynchronous Environments},
  author={Froger, Romain and Andrews, Pierre and Bettini, Matteo and Budhiraja, Amar and Cabral, Ricardo Silveira and Do, Virginie and Garreau, Emilien and Gaya, Jean-Baptiste and Lauren{\c{c}}on, Hugo and Lecanu, Maxime and others},
  journal={arXiv preprint arXiv:2602.11964},
  year={2026}
}

@article{lai2025mini,
  title={Mini-o3: Scaling up reasoning patterns and interaction turns for visual search},
  author={Lai, Xin and Li, Junyi and Li, Wei and Liu, Tao and Li, Tianjian and Zhao, Hengshuang},
  journal={arXiv preprint arXiv:2509.07969},
  year={2025}
}

@article{shen2023hugginggpt,
  title={Hugginggpt: Solving ai tasks with chatgpt and its friends in hugging face},
  author={Shen, Yongliang and Song, Kaitao and Tan, Xu and Li, Dongsheng and Lu, Weiming and Zhuang, Yueting},
  journal={Advances in Neural Information Processing Systems},
  volume={36},
  pages={38154--38180},
  year={2023}
}

@article{deng2023mind2web,
  title={Mind2web: Towards a generalist agent for the web},
  author={Deng, Xiang and Gu, Yu and Zheng, Boyuan and Chen, Shijie and Stevens, Sam and Wang, Boshi and Sun, Huan and Su, Yu},
  journal={Advances in Neural Information Processing Systems},
  volume={36},
  pages={28091--28114},
  year={2023}
}

@article{zeng2026vision,
  title={Vision-DeepResearch Benchmark: Rethinking Visual and Textual Search for Multimodal Large Language Models},
  author={Zeng, Yu and Huang, Wenxuan and Fang, Zhen and Chen, Shuang and Shen, Yufan and Cai, Yishuo and Wang, Xiaoman and Yin, Zhenfei and Chen, Lin and Chen, Zehui and others},
  journal={arXiv preprint arXiv:2602.02185},
  year={2026}
}

@article{narayan2025deepmmsearch,
  title={Deepmmsearch-r1: Empowering multimodal llms in multimodal web search},
  author={Narayan, Kartik and Xu, Yang and Cao, Tian and Nerella, Kavya and Patel, Vishal M and Shiee, Navid and Grasch, Peter and Jia, Chao and Yang, Yinfei and Gan, Zhe},
  journal={arXiv preprint arXiv:2510.12801},
  year={2025}
}

@inproceedings{yue2024mmmu,
  title={Mmmu: A massive multi-discipline multimodal understanding and reasoning benchmark for expert agi},
  author={Yue, Xiang and Ni, Yuansheng and Zhang, Kai and Zheng, Tianyu and Liu, Ruoqi and Zhang, Ge and Stevens, Samuel and Jiang, Dongfu and Ren, Weiming and Sun, Yuxuan and others},
  booktitle={Proceedings of the IEEE/CVF conference on computer vision and pattern recognition},
  pages={9556--9567},
  year={2024}
}

@article{yu2023mm,
  title={Mm-vet: Evaluating large multimodal models for integrated capabilities},
  author={Yu, Weihao and Yang, Zhengyuan and Li, Linjie and Wang, Jianfeng and Lin, Kevin and Liu, Zicheng and Wang, Xinchao and Wang, Lijuan},
  journal={arXiv preprint arXiv:2308.02490},
  year={2023}
}

@article{huang2026vision,
  title={Vision-DeepResearch: Incentivizing DeepResearch Capability in Multimodal Large Language Models},
  author={Huang, Wenxuan and Zeng, Yu and Wang, Qiuchen and Fang, Zhen and Cao, Shaosheng and Chu, Zheng and Yin, Qingyu and Chen, Shuang and Yin, Zhenfei and Chen, Lin and others},
  journal={arXiv preprint arXiv:2601.22060},
  year={2026}
}

@article{huang2026mmdeepresearch,
  title={MMDeepResearch-Bench: A Benchmark for Multimodal Deep Research Agents},
  author={Huang, Peizhou and Zhong, Zixuan and Wan, Zhongwei and Zhou, Donghao and Alam, Samiul and Wang, Xin and Li, Zexin and Dou, Zhihao and Zhu, Li and Xiong, Jing and others},
  journal={arXiv preprint arXiv:2601.12346},
  year={2026}
}

@article{bai2025qwen3,
  title={Qwen3-vl technical report},
  author={Bai, Shuai and Cai, Yuxuan and Chen, Ruizhe and Chen, Keqin and Chen, Xionghui and Cheng, Zesen and Deng, Lianghao and Ding, Wei and Gao, Chang and Ge, Chunjiang and others},
  journal={arXiv preprint arXiv:2511.21631},
  year={2025}
}

@article{team2026kimi,
  title={Kimi K2. 5: Visual Agentic Intelligence},
  author={Team, Kimi and Bai, Tongtong and Bai, Yifan and Bao, Yiping and Cai, SH and Cao, Yuan and Charles, Y and Che, HS and Chen, Cheng and Chen, Guanduo and others},
  journal={arXiv preprint arXiv:2602.02276},
  year={2026}
}

@misc{qwen35blog,
    title = {Qwen3.5: Accelerating Productivity with Native Multimodal Agents},
    url = {https://qwen.ai/blog?id=qwen3.5},
    author = {Qwen Team},
    month = {February},
    year = {2026}
}

@article{hurst2024gpt,
  title={Gpt-4o system card},
  author={Hurst, Aaron and Lerer, Adam and Goucher, Adam P and Perelman, Adam and Ramesh, Aditya and Clark, Aidan and Ostrow, AJ and Welihinda, Akila and Hayes, Alan and Radford, Alec and others},
  journal={arXiv preprint arXiv:2410.21276},
  year={2024}
}

@inproceedings{mavors,
  title={Mavors: Multi-granularity video representation for multimodal large language model},
  author={Shi, Yang and Liu, Jiaheng and Guan, Yushuo and Wu, Zhenhua and Zhang, Yuanxing and Wang, Zihao and Lin, Weihong and Hua, Jingyun and Wang, Zekun and Chen, Xinlong and others},
  booktitle={Proceedings of the 33rd ACM International Conference on Multimedia},
  pages={10994--11003},
  year={2025}
}

@article{monet,
  title={Monet: Reasoning in latent visual space beyond images and language},
  author={Wang, Qixun and Shi, Yang and Wang, Yifei and Zhang, Yuanxing and Wan, Pengfei and Gai, Kun and Ying, Xianghua and Wang, Yisen},
  journal={arXiv preprint arXiv:2511.21395},
  year={2025}
}

@article{mmevideoocr,
  title={Mme-videoocr: Evaluating ocr-based capabilities of multimodal llms in video scenarios},
  author={Shi, Yang and Wang, Huanqian and Xie, Wulin and Zhang, Huanyao and Zhao, Lijie and Zhang, Yi-Fan and Li, Xinfeng and Fu, Chaoyou and Wen, Zhuoer and Liu, Wenting and others},
  journal={arXiv preprint arXiv:2505.21333},
  year={2025}
}

@article{realunify,
  title={Realunify: Do unified models truly benefit from unification? a comprehensive benchmark},
  author={Shi, Yang and Dong, Yuhao and Ding, Yue and Wang, Yuran and Zhu, Xuanyu and Zhou, Sheng and Liu, Wenting and Tian, Haochen and Wang, Rundong and Wang, Huanqian and others},
  journal={arXiv preprint arXiv:2509.24897},
  year={2025}
}

\clearpage
\appendix
\renewcommand{\theHsection}{\Alph{section}}
\renewcommand{\theHtable}{\Alph{section}.\arabic{table}}
\renewcommand{\theHfigure}{\Alph{section}.\arabic{figure}}
\begin{center}
    \Large\bfseries Supplementary Material
\end{center}
\vspace{1em}

\section{Process-Aware Evaluation Protocol and Scoring Details}
\label{app:eval_protocol}

A core contribution of is enabling automated, process-level evaluation without human intervention. We score agents using final-answer correctness and step-wise checkpoint compliance along two orthogonal axes: Strategy (\textbf{S-axis}) and Visual Evidence (\textbf{V-axis}). Checkpoint matching relies strictly on the logged execution traces and generated artifacts.

\subsection{S-axis Evaluation (Strategy \& Tool Execution)}
S-axis checkpoints automatically verify whether the agent adopted the correct high-level plan and invoked the necessary tools. To robustly handle heterogeneous agent architectures, we apply distinct parsing strategies:
\begin{itemize}[leftmargin=*,itemsep=0pt,parsep=0pt,topsep=2pt]
    \item \textit{Pre-defined Tool Agents:} We directly parse the structured function calls to compare the tool selection and arguments against the reference strategy.
    \item \textit{Code-Writing Agents:} For models that write free-form code, we perform Abstract Syntax Tree (AST) parsing on the executed Python snippets to extract canonical visual operations, ensuring the semantic intent matches the ground truth regardless of coding style.
    \item \textit{Retrieval Actions:} Web search results are inherently non-deterministic, making exact string matching insufficient. Instead, we deploy an external LLM-as-a-judge. We provide the judge with the annotated step \emph{description}, \emph{search keywords}, \emph{expected answer}, and the agent's \emph{returned observation} (API payload). The S-axis checkpoint passes if the judge determines the returned payload contains sufficient information to satisfy the expected intermediate answer.
\end{itemize}

\subsection{V-axis Evaluation (Visual Evidence Verification)}
Even if an agent uses the correct tool, it may extract the wrong region (e.g., cropping the background instead of the target). Inspired by the visual faithfulness evaluation in CodeV~\citep{hou2025codev}, the V-axis explicitly verifies whether the intermediate visual artifacts \emph{genuinely contain} the decisive evidence.

Each V-axis checkpoint is paired with a highly specific intermediate question (e.g., ``Does this cropped image clearly show the red mailbox?''). During execution, a single code block or tool call may yield multiple sub-images. We submit \emph{all} produced visual artifacts to an external VLM judge (e.g., GPT-4o) alongside the intermediate question. 

Crucially, because all intermediate outputs are appended to the agent's multimodal context history for subsequent turns, the agent successfully ``sees'' the evidence if at least one artifact is correct. Therefore, we apply an \textbf{any-pass mechanism}: the V-axis checkpoint is marked as passed if \emph{any} of the processed sub-images passes the VLM judge's evaluation.

\subsection{Final Scores and Efficiency (Overthink)}
Let $\mathcal{C}_S$ and $\mathcal{C}_V$ denote the strategy and visual checkpoint sets for a task. The process scores are the fractions of passed checkpoints:
\begin{equation}
    S = \frac{1}{|\mathcal{C}_S|}\sum_{c\in\mathcal{C}_S}\mathbb{I}[c\ \text{passed}], \quad V = \frac{1}{|\mathcal{C}_V|}\sum_{c\in\mathcal{C}_V}\mathbb{I}[c\ \text{passed}]
\end{equation}
Final answer accuracy (\textbf{Acc}) is computed via normalized exact match.

To quantify redundant interactions, we compute an \textbf{Overthink} penalty relative to the minimal human reference trajectory:
\begin{equation}
    \text{Overthink}=\frac{\max(0, C_{\text{agent}}-C_{\text{human}})}{C_{\text{human}}+1}
\end{equation}
where $C_{\text{agent}}$ and $C_{\text{human}}$ count the interactions (tool calls or executed code blocks) that produce new observable artifacts. All evaluations run under a strict interaction budget to prevent degenerate looping.

\subsection{Judge Prompt Templates}
\label{app:judge_prompts}

To make the evaluation fully reproducible, we release the exact prompt templates used by the automated judges. In our implementation, the two axes are scored with different prompts because they inspect different evidence modalities.

\paragraph{S-axis judge}
For search-related checkpoints, the judge receives the checkpoint description, the model's actual search queries, and a compact summary of the returned search results. The prompt is intentionally \emph{lenient} on query wording but \emph{strict} on whether the expected information was actually retrieved.

\begin{tcolorbox}[
    enhanced,
    breakable,
    colframe=black!70,
    colback=yellow!5,
    boxrule=1pt, arc=4mm,
    left=2mm, right=2mm, top=1mm, bottom=1mm,
    title={S-axis judge prompt template}
]
\small
\begin{verbatim}
You are evaluating whether a model's search successfully found the
required information.

**Expected Search Strategy:**
{checkpoint_desc}

**CRITICAL: Expected Answer to Find:**
{expected_answer}   # included only when available

**Model's Actual Search Queries:**
{queries_text}

**Search Results Summary:**
{results_text}

**Evaluation Criteria (Two-Stage Check):**

STAGE 1 - Query Relevance (Lenient Requirement):
- Be LENIENT when evaluating query quality - broad queries are
  acceptable if they're reasonable
- Accept queries that mention the main entity, even if not specific
- Accept synonyms, paraphrases, different word orders, translations
- Reject ONLY IF: wrong entity OR completely unrelated topic

STAGE 2 - Expected Answer (Primary Criterion):
- This is the MAIN criterion for passing
- Do the search results actually contain the answer/information?
- If an expected answer is specified, it appears in the search results

**Decision Logic:**
1. If queries are completely unrelated to the topic -> FAIL
2. If queries are reasonable but 
   results don't contain expected answer -> FAIL
3. If queries are reasonable and results contain 
   expected answer -> PASS

**Response Format:**
VERDICT: [PASS/FAIL]
REASONING: [brief explanation]
\end{verbatim}
\end{tcolorbox}

\paragraph{V-axis judge (for \textbf{$V_{true}$} ).}
For \textbf{$V_{true}$} checkpoints, the judge is not asked to reason about the whole task. Instead, it receives a single intermediate artifact and a highly specific question designed by the annotator (e.g., ``What is the road name shown in this crop?''). In the released implementation, the judge runs independently on \emph{each} produced artifact, and the checkpoint passes under an any-pass rule if any artifact contains the required evidence.

\begin{tcolorbox}[
    enhanced,
    breakable,
    colframe=black!70,
    colback=yellow!5,
    boxrule=1pt, arc=4mm,
    left=2mm, right=2mm, top=1mm, bottom=1mm,
    title={V-axis judge prompt template}
]
\small
\begin{verbatim}
System:
You are a helpful assistant that answers questions about images.
Provide concise, accurate answers based on what you see in the image.

User input:
[intermediate visual artifact]
{visual_question}

Judging rule applied by the harness:
- Run the judge on every generated intermediate image.
- Compare the returned answer against the expected short answer.
- Mark the checkpoint as passed if ANY image returns the answer.
\end{verbatim}
\end{tcolorbox}

\paragraph{Judge decoding setup.}
\section{Tool API Specification and Implementation} 
\label{app:tools} 

\subsection{Overview} 

\begin{tcolorbox}[ 
    enhanced, 
    breakable, 
    colframe=black!70, 
    colback=yellow!5, 
    boxrule=1pt, arc=4mm, 
    left=2mm, right=2mm, top=1mm, bottom=1mm, 
    before=\needspace{6\baselineskip}, 
    title={Overview} 
] 
\noindent\textbf{Goal.} This appendix documents the exact tool API used by Agentic-MME, including the \emph{raw JSON schema} consumed by OpenAI-compatible tool calling. All intermediate artifacts are tracked by \texttt{image\_index}: Image~0 is the original input, and every successful image tool call appends one or several new images and returns \texttt{new\_image\_index}. 
\end{tcolorbox}

\subsection{Canonical Function-Calling Interface (OpenAI-Compatible)} 

\begin{tcolorbox}[ 
    enhanced, 
    breakable, 
    colframe=black!70, 
    colback=yellow!5, 
    boxrule=1pt, arc=4mm, 
    left=2mm, right=2mm, top=1mm, bottom=1mm, 
    before=\needspace{8\baselineskip}, 
    title={Example tool call payload (OpenAI-compatible wrapper)} 
] 
\begin{verbatim}
A tool call is identified by function 
name and a JSON object function.arguments
{ 
  "type": "function", 
  "function": { 
    "name": "<tool_name>", 
    "arguments": "{... JSON string ...}" 
  } 
} 
\end{verbatim}
\end{tcolorbox}

\subsection{Atomic Image Tools (Visual Expansion)}

\begin{tcolorbox}[
    enhanced,
    breakable,
    colframe=black!70,
    colback=yellow!5,
    boxrule=1pt, arc=4mm,
    left=2mm, right=2mm, top=1mm, bottom=1mm,
    before=\needspace{8\baselineskip},
    title={OpenAI-compatible schema: Geometric \& Spatial Tools}
]
\begin{verbatim}
[
  {
    "type": "function",
    "function": {
      "name": "crop",
      "description": "Crop/zoom into a region of an image using
      normalized bounding box coordinates (0-1000 scale).",
      "parameters": {
        "type": "object",
        "properties": {
          "image_index": {
            "type": "integer",
            "description": "Index of the image to operate on
            (0 = original, 1, 2... = processed images)",
            "minimum": 0
          },
          "bbox_2d": {
            "type": "array",
            "items": {
              "type": "number"
            },
            "minItems": 4,
            "maxItems": 4,
            "description": "Bounding box [x1, y1, x2, y2]
            in 0-1000 normalized coordinates"
          },
          "zoom_scale": {
            "type": "number",
            "description": "Magnification factor for the
            cropped region (default: 1.0). Higher values
            produce larger images.",
            "minimum": 0.5,
            "maximum": 5.0,
            "default": 1.0
          },
          "label": {
            "type": "string",
            "description": "Description of what this operation
                            is for"
          }
        },
        "required": [
          "image_index",
          "bbox_2d"
        ]
      }
    }
  },
  {
    "type": "function",
    "function": {
      "name": "rotate",
      "description": "Rotate an image by angle degrees
      (positive = counterclockwise).",
      "parameters": {
        "type": "object",
        "properties": {
          "image_index": {
            "type": "integer",
            "description": "Index of the image to operate on
            (0 = original, 1, 2... = processed images)",
            "minimum": 0
          },
          "angle": {
            "type": "number",
            "description": "Rotation angle in degrees"
          },
          "expand": {
            "type": "boolean",
            "description": "If true, expand canvas to fit
            rotated image. Default: true"
          },
          "label": {
            "type": "string",
            "description": "Description of what this operation
                            is for"
          }
        },
        "required": [
          "image_index",
          "angle"
        ]
      }
    }
  },
  {
    "type": "function",
    "function": {
      "name": "flip",
      "description": "Flip/mirror an image horizontally,
      vertically, or both directions.",
      "parameters": {
        "type": "object",
        "properties": {
          "image_index": {
            "type": "integer",
            "description": "Index of the image to operate on
            (0 = original, 1, 2... = processed images)",
            "minimum": 0
          },
          "direction": {
            "type": "string",
            "enum": [
              "horizontal",
              "vertical",
              "both"
            ],
            "description": "Flip direction: 'horizontal'
            (left-right mirror), 'vertical' (top-bottom
             mirror), or 'both'. Default: horizontal"
          },
          "label": {
            "type": "string",
            "description": "Description of what this operation
                            is for"
          }
        },
        "required": [
          "image_index"
        ]
      }
    }
  },
  {
    "type": "function",
    "function": {
      "name": "resize",
      "description": "Resize an image to (width, height)
      OR by scale factor.",
      "parameters": {
        "type": "object",
        "properties": {
          "image_index": {
            "type": "integer",
            "description": "Index of the image to operate on
            (0 = original, 1, 2... = processed images)",
            "minimum": 0
          },
          "width": {
            "type": "integer",
            "description": "Target width in pixels"
          },
          "height": {
            "type": "integer",
            "description": "Target height in pixels"
          },
          "scale": {
            "type": "number",
            "description": "Scale factor
            (e.g., 2.0 = double size, 0.5 = half)"
          },
          "label": {
            "type": "string",
            "description": "Description of what this operation
                            is for"
          }
        },
        "required": [
          "image_index"
        ]
      }
    }
  }
]
\end{verbatim}
\end{tcolorbox}

\begin{tcolorbox}[
    enhanced,
    breakable,
    colframe=black!70,
    colback=yellow!5,
    boxrule=1pt, arc=4mm,
    left=2mm, right=2mm, top=1mm, bottom=1mm,
    before=\needspace{8\baselineskip},
    title={OpenAI-compatible schema: Color \& Contrast Tools}
]
\begin{verbatim}
[
  {
    "type": "function",
    "function": {
      "name": "grayscale",
      "description": "Convert an image to grayscale.",
      "parameters": {
        "type": "object",
        "properties": {
          "image_index": {
            "type": "integer",
            "description": "Index of the image to operate on
            (0 = original, 1, 2... = processed images)",
            "minimum": 0
          },
          "label": {
            "type": "string",
            "description": "Description of what this operation
                            is for"
          }
        },
        "required": [
          "image_index"
        ]
      }
    }
  },
  {
    "type": "function",
    "function": {
      "name": "autocontrast",
      "description": "Apply automatic contrast adjustment.",
      "parameters": {
        "type": "object",
        "properties": {
          "image_index": {
            "type": "integer",
            "description": "Index of the image to operate on
            (0 = original, 1, 2... = processed images)",
            "minimum": 0
          },
          "cutoff": {
            "type": "number",
            "description": "Percentage of lightest/darkest
            pixels to ignore. Default: 0"
          },
          "label": {
            "type": "string",
            "description": "Description of what this operation
                            is for"
          }
        },
        "required": [
          "image_index"
        ]
      }
    }
  },
  {
    "type": "function",
    "function": {
      "name": "invert",
      "description": "Invert the colors of an image.",
      "parameters": {
        "type": "object",
        "properties": {
          "image_index": {
            "type": "integer",
            "description": "Index of the image to operate on
            (0 = original, 1, 2... = processed images)",
            "minimum": 0
          },
          "label": {
            "type": "string",
            "description": "Description of what this operation
                            is for"
          }
        },
        "required": [
          "image_index"
        ]
      }
    }
  },
  {
    "type": "function",
    "function": {
      "name": "equalize",
      "description": "Equalize histogram for better
                      contrast distribution.",
      "parameters": {
        "type": "object",
        "properties": {
          "image_index": {
            "type": "integer",
            "description": "Index of the image to operate on
            (0 = original, 1, 2... = processed images)",
            "minimum": 0
          },
          "label": {
            "type": "string",
            "description": "Description of what this operation
                            is for"
          }
        },
        "required": [
          "image_index"
        ]
      }
    }
  },
  {
    "type": "function",
    "function": {
      "name": "threshold",
      "description": Apply threshold to convert image to binary
      "parameters": {
        "type": "object",
        "properties": {
          "image_index": {
            "type": "integer",
            "description": "Index of the image to operate on
            (0 = original, 1, 2... = processed images)",
            "minimum": 0
          },
          "value": {
            "type": "integer",
            "description": "Threshold value (0-255)"
          },
          "mode": {
            "type": "string",
            "enum": [
              "binary",
              "binary_inv",
              "trunc",
              "tozero"
            ],
            "description": "Threshold mode. Default: binary"
          },
          "label": {
            "type": "string",
            "description": "Description of what this operation
                            is for"
          }
        },
        "required": [
          "image_index"
        ]
      }
    }
  }
]
\end{verbatim}
\end{tcolorbox}

\begin{tcolorbox}[
    enhanced,
    breakable,
    colframe=black!70,
    colback=yellow!5,
    boxrule=1pt, arc=4mm,
    left=2mm, right=2mm, top=1mm, bottom=1mm,
    before=\needspace{8\baselineskip},
    title={OpenAI-compatible schema: Filtering \& Advanced Tools}
]
\begin{verbatim}
[
  {
    "type": "function",
    "function": {
      "name": "blur",
      "description": "Apply Gaussian blur.",
      "parameters": {
        "type": "object",
        "properties": {
          "image_index": {
            "type": "integer",
            "description": "Index of the image to operate on
            (0 = original, 1, 2... = processed images)",
            "minimum": 0
          },
          "radius": {
            "type": "integer",
            "description": "Blur radius. Default: 2"
          },
          "label": {
            "type": "string",
            "description": "Description of what this operation
                            is for"
          }
        },
        "required": [
          "image_index"
        ]
      }
    }
  },
  {
    "type": "function",
    "function": {
      "name": "sharpen",
      "description": "Apply sharpening filter.",
      "parameters": {
        "type": "object",
        "properties": {
          "image_index": {
            "type": "integer",
            "description": "Index of the image to operate on
            (0 = original, 1, 2... = processed images)",
            "minimum": 0
          },
          "label": {
            "type": "string",
            "description": "Description of what this operation
                            is for"
          }
        },
        "required": [
          "image_index"
        ]
      }
    }
  },
  {
    "type": "function",
    "function": {
      "name": "denoise",
      "description": "Remove noise from an image.",
      "parameters": {
        "type": "object",
        "properties": {
          "image_index": {
            "type": "integer",
            "description": "Index of the image to operate on
            (0 = original, 1, 2... = processed images)",
            "minimum": 0
          },
          "strength": {
            "type": "integer",
            "description": "Denoising strength (1-30)."
          },
          "label": {
            "type": "string",
            "description": "Description of what this operation
                            is for"
          }
        },
        "required": [
          "image_index"
        ]
      }
    }
  },
  {
    "type": "function",
    "function": {
      "name": "edge_detect",
      "description": "Detect edges in an image.",
      "parameters": {
        "type": "object",
        "properties": {
          "image_index": {
            "type": "integer",
            "description": "Index of the image to operate on
            (0 = original, 1, 2... = processed images)",
            "minimum": 0
          },
          "method": {
            "type": "string",
            "enum": [
              "canny",
              "sobel",
              "simple"
            ],
            "description": "Edge detection method.
             Default: canny"
          },
          "label": {
            "type": "string",
            "description": "Description of what this operation
                            is for"
          }
        },
        "required": [
          "image_index"]}}
          } 
        ]

\end{verbatim}
\end{tcolorbox}

\subsection{Web Retrieval Tools (Knowledge Expansion)}
\label{app:web_tools}

Beyond the visual operations, Agentic-MME exposes four retrieval tools for \emph{Knowledge Expansion}. The released implementation uses Serper.dev for Google text search and Google Lens, Jina Reader for webpage fetching, and an auxiliary image-downloading tool that allows the agent to bring externally retrieved candidate images back into the local image workspace for further inspection. 

\begin{itemize}[leftmargin=*,itemsep=-0.1em,topsep=0.1em]
  \renewcommand\labelitemi{$\diamond$}  
    \item \textbf{google\_search}: Google text search via Serper.dev, returning top results.
    \item \textbf{google\_lens\_search}: reverse image search via Serper.dev Lens. If the target is a local image file, the harness first uploads the image to a temporary image host and then queries Lens using the resulting public URL.
    \item \textbf{fetch\_webpage}: webpage-to-text extraction via Jina Reader.
    \item \textbf{download\_image}: downloads a result image URL into the sandbox.
    \item \textbf{Caching and replay}: all search requests can be stored as per-task JSON payloads to support deterministic replay; when replay mode is enabled, the harness reuses cached responses instead of making fresh web requests.
\end{itemize}

\begin{tcolorbox}[
    enhanced,
    breakable,
    colframe=black!70,
    colback=yellow!5,
    boxrule=1pt, arc=4mm,
    left=2mm, right=2mm, top=1mm, bottom=1mm,
    title={OpenAI-compatible schema: Web Retrieval Tools}
]
\begin{verbatim}
[
  {
    "type": "function",
    "function": {
      "name": "google_search",
      "description": "Search the web using Google via Serper  
      API.Use for facts, current information, specifications,
       prices, or any knowledge queries.",
      "parameters": {
        "type": "object",
        "properties": {
          "query": {
            "type": "string",
            "description": "The search query."
          },
          "gl": {
            "type": "string",
            "description": "Geo location code (e.g., 'us').
            Default: 'us'"
          },
          "hl": {
            "type": "string",
            "description": "Language code (e.g., 'en', 'zh').
            Default: 'en'"
          }
        },
        "required": ["query"]
      }
    }
  },
  {
    "type": "function",
    "function": {
      "name": "google_lens_search",
      "description": "Reverse image search using Google Lens
       via Serper.dev API. Use to identify objects,brands,
        logos, landmarks, products, or text in images.",
      "parameters": {
        "type": "object",
        "properties": {
          "image_ref": {
            "type": "string",
            "enum": ["current", "original"],
            "description": "Quick reference: 'current' for the
             latest processed image, 'original' for the input"
          },
          "image_path": {
            "type": "string",
            "description": "Filename or full path to a
             specific image.After image processing, the 
             agent may use a filename such as
            'transformed_image_1.png' to search that artifact."
          }
        },
        "required": []
      }
    }
  },
  {
    "type": "function",
    "function": {
      "name": "fetch_webpage",
      "description": "Fetch and read the content of a webpage.
      Returns clean text extracted from the URL via Jina Reader.
      "parameters": {
        "type": "object",
        "properties": {
          "url": {
            "type": "string",
            "description": "The webpage URL to fetch 
             (must be http/https)."
          },
          "max_chars": {
            "type": "integer",
            "description": "Maximum characters to return.
             Default: 12000"
          }
        },
        "required": ["url"]
      }
    }
  },
  {
    "type": "function",
    "function": {
      "name": "download_image",
      "description": "Download an image from a URL into the 
       local sandbox. Downloaded images are appended to the 
       image list and can be used by the same visual tools as 
        ordinary inputs. For safety and cost control, 
        the harness caps downloads per task.",
      "parameters": {
        "type": "object",
        "properties": {
          "url": {
            "type": "string",
            "description": "Direct image URL to download."
          }
        },
        "required": ["url"]
      }
    }
  }
]
\end{verbatim}
\end{tcolorbox}

\section{More Dataset Statistics}
\begin{figure}[htbp]
    \centering
    \includegraphics[width=0.6\linewidth]{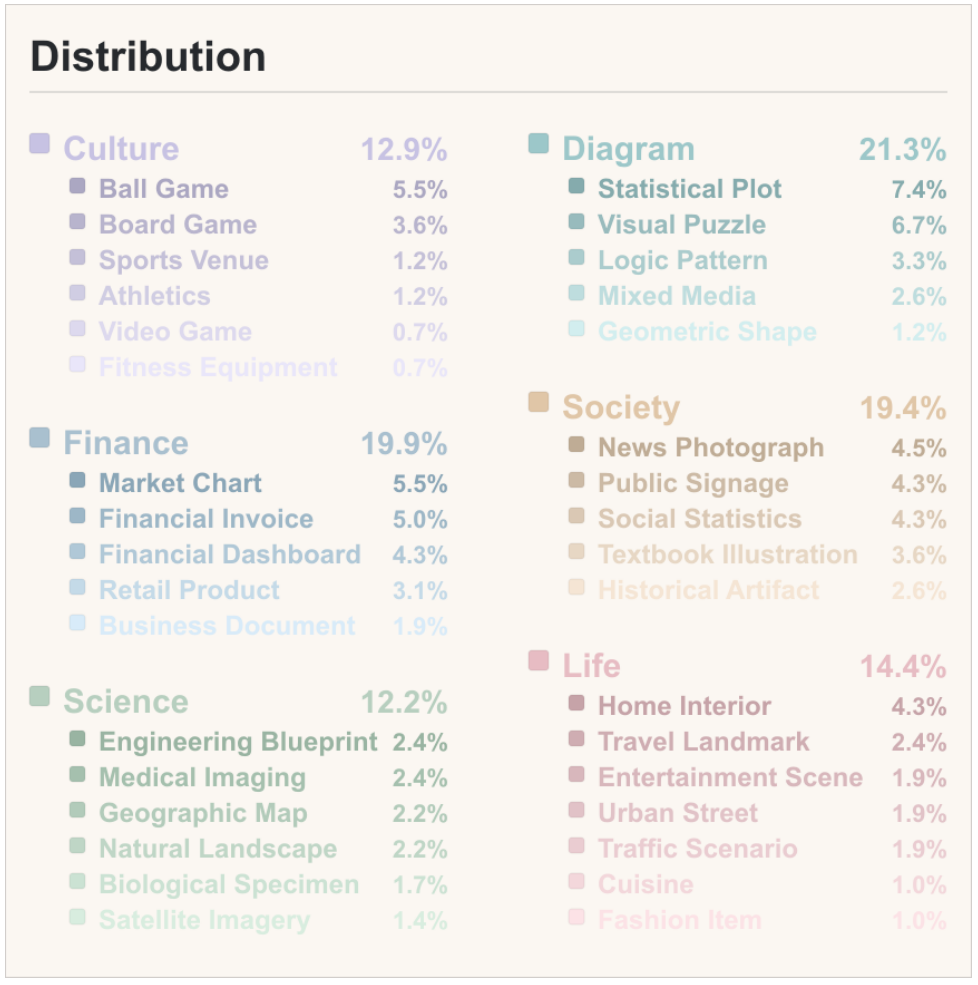}
    \caption{\textbf{more fine-grained domain distribution}}
    \label{fig:dataset_dim}
\end{figure}
\section{Case Studies}
\label{app:cases}
This appendix presents representative cases for the levels defined in Sec.~\ref{sec:task_setup}. 
\subsection{Level 1: Core Visual Expansion}
Level~1 tasks isolate the core perception--action loop of Visual Expansion. 
As illustrated in Figures~\ref{fig:app_case_l1_text} and~\ref{fig:app_case_l1_person}, these tasks are dominated by a single decisive visual operation that surfaces otherwise inaccessible evidence. 
They do not require multi-step search, cross-image bookkeeping, or iterative hypothesis refinement.

\begin{center}
    \includegraphics[width=0.88\linewidth]{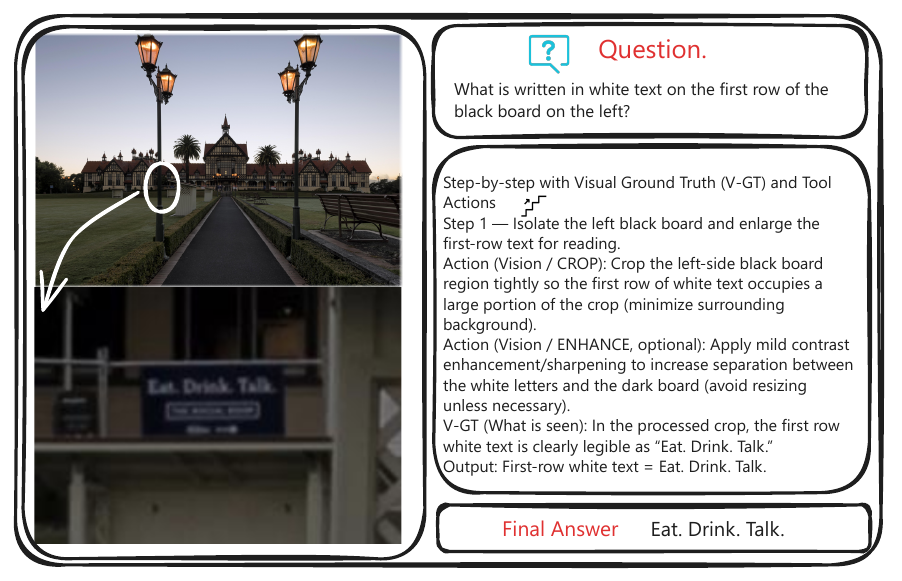}
    \captionof{figure}{\textbf{Level 1 Case: Text Extraction via Localized Enhancement.} 
    The decisive cue is a small blurry background text region that must be isolated and enhanced before it becomes reliably readable.}
    \label{fig:app_case_l1_text}
\end{center}

\noindent\textbf{Figure~\ref{fig:app_case_l1_text}.}
This case represents a typical Level~1 workflow: the answer-bearing signal is already present in the original image, but passive perception is insufficient because the relevant text is too small or too blurry. 
A single localized crop, optionally followed by enhancement, is enough to reveal the answer. 
The task therefore tests whether the model can proactively surface the correct local evidence rather than relying on one-shot visual guessing.

\vspace{1mm}

\begin{center}
    \includegraphics[width=0.68\linewidth]{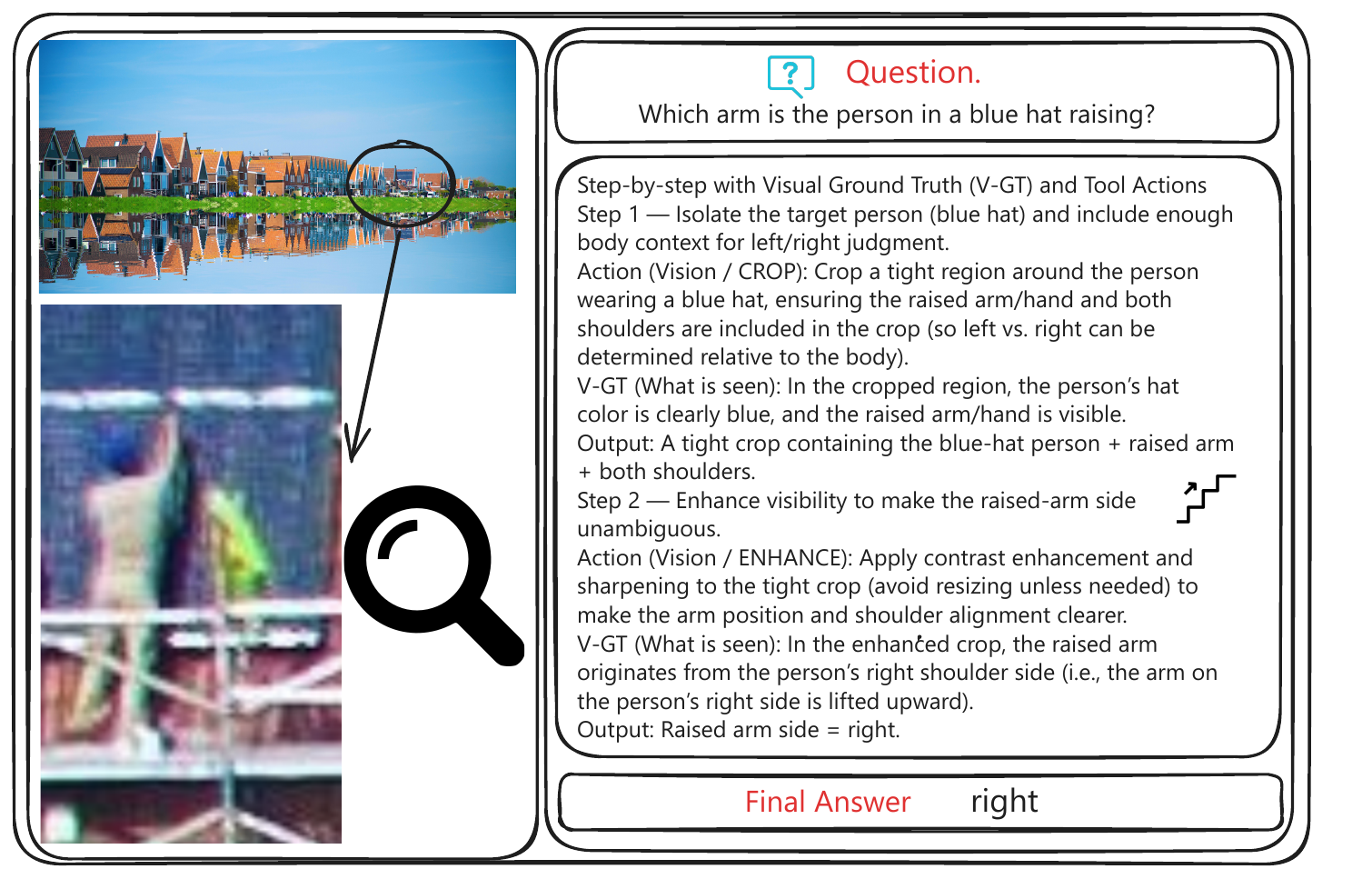}
    \captionof{figure}{\textbf{Level 1 Case: Fine-grained Detail Isolation.} 
    The agent must tightly isolate a distant human figure to resolve a subtle spatial attribute that is ambiguous in the raw image.}
    \label{fig:app_case_l1_person}
\end{center}

\noindent\textbf{Figure~\ref{fig:app_case_l1_person}.}
This example also falls squarely into Level~1. 
The core challenge is to isolate a small person region and recover a fine-grained spatial cue, such as the configuration of the arms, that cannot be read reliably from the full image. 
Unlike higher-level cases, the workflow remains a single-step visual evidence recovery episode and does not require external knowledge or multi-stage planning.

\subsection{Level 2: Short Multi-Step Workflows}

Level~2 tasks require more than one action, but the solution path is still short and mostly linear. 
Figures~\ref{fig:app_case_l2_clipboard} and~\ref{fig:app_case_l2_flight} illustrate two representative Level~2 patterns: short-horizon multi-region composition and structured multi-step visual bookkeeping. 
Compared with Level~1, the difficulty comes from coordinating multiple intermediate visual observations rather than from a single decisive crop.

\begin{center}
    \includegraphics[width=0.88\linewidth]{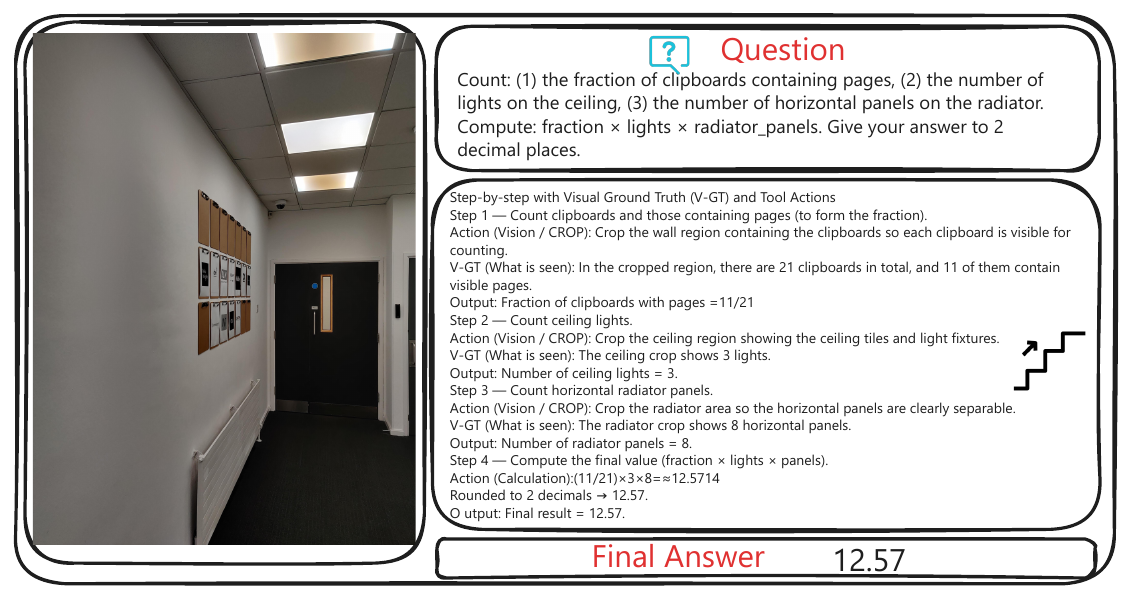}
    \captionof{figure}{\textbf{Level 2 Case: Multi-Region Counting and Arithmetic Composition.}}
    \label{fig:app_case_l2_clipboard}
\end{center}

\noindent\textbf{Figure~\ref{fig:app_case_l2_clipboard}.}
This case is harder than Level~1 because the answer is not tied to one isolated crop. 
Instead, the model must perform several localized inspections, preserve multiple intermediate counts, and compose them correctly at the end. 
However, the workflow is still fundamentally linear: once the relevant regions are identified, the remaining difficulty lies in short-horizon aggregation rather than in iterative search--vision interaction. 
\vspace{1mm}

\begin{center}
    \includegraphics[width=0.68\linewidth]{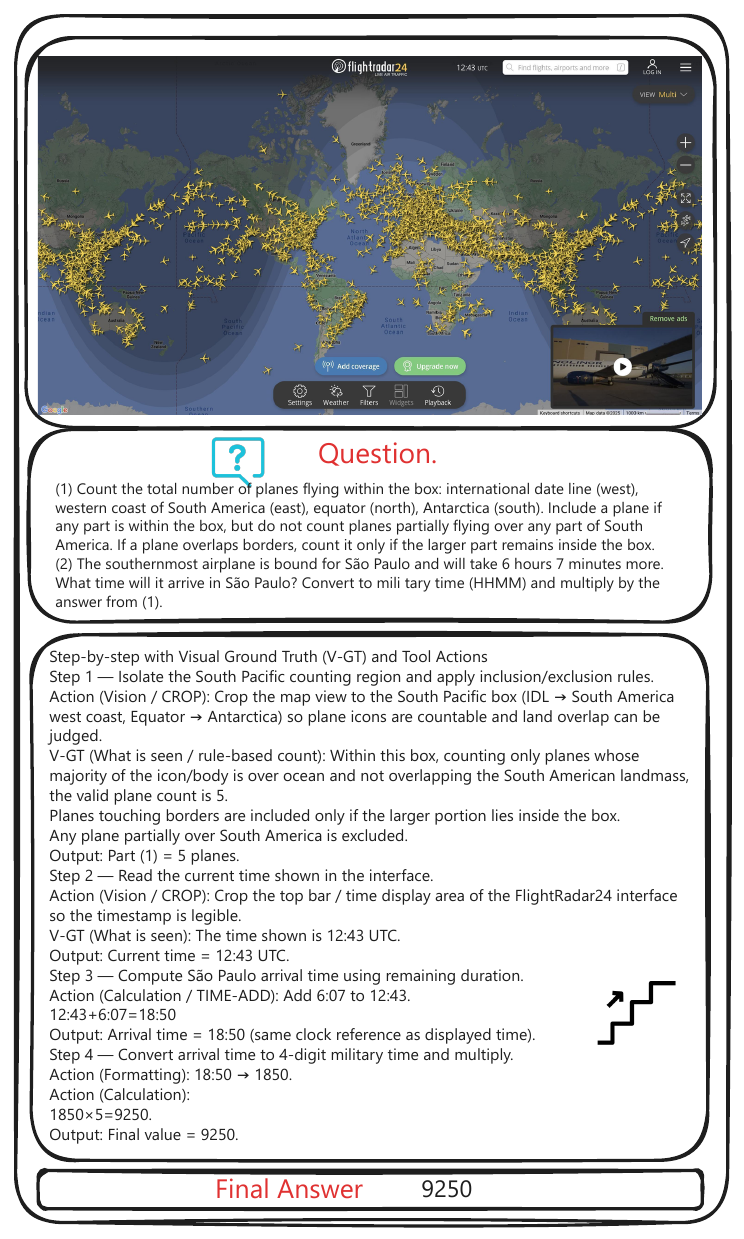}
    \captionof{figure}{\textbf{Level 2 Case: Region-Constrained Counting with Temporal Arithmetic.}
    The agent isolates the South Pacific region in the flight map, applies inclusion/exclusion counting rules, then reads a separate time cue and combines the two intermediate results through temporal arithmetic.}
    \label{fig:app_case_l2_flight}
\end{center}

\noindent\textbf{Figure~\ref{fig:app_case_l2_flight}.}
This example represents another Level~2 pattern in which the model must complete a short sequence of heterogeneous but still straightforward subgoals: region selection, constrained counting, time reading, and final formatting. 
The task requires explicit intermediate bookkeeping, but it does not yet demand repeated external verification or deep interleaving between visual reasoning and search. 
It therefore sits naturally in the middle layer between Level~1 single-step perception and Level~3 synergistic workflows.

\subsection{Level 3: Advanced Synergistic Problem Solving}

Level~3 cases go beyond short sequential chaining. 
As shown in Figures~\ref{fig:app_case_l3_map}, \ref{fig:app_case_l3_tile}, \ref{fig:app_case_l3_gas}, and~\ref{fig:app_case_l3_hockey}, these tasks require either advanced global image analysis or a truly intertwined interaction between localized visual evidence and external knowledge. 
In these settings, the model must refine hypotheses across multiple stages rather than simply execute a fixed linear sequence.

\begin{center}
    \includegraphics[width=0.88\linewidth]{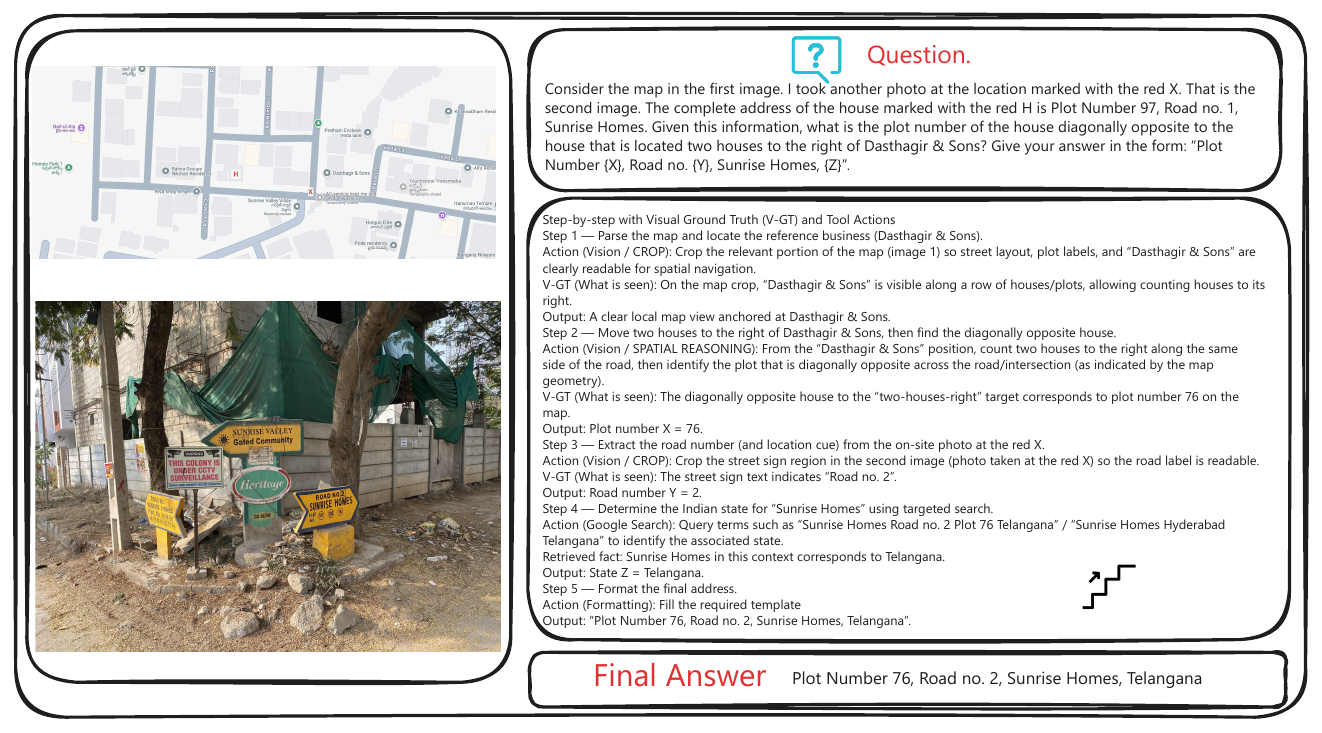}
    \captionof{figure}{\textbf{Level 3 Case: Cross-Image Spatial Grounding with Knowledge Verification.}
    The agent first crops the map to anchor on the reference business and perform local spatial reasoning, then crops the companion street-view image to recover the road cue, and finally uses open-web retrieval to determine the associated state.}
    \label{fig:app_case_l3_map}
\end{center}

\noindent\textbf{Figure~\ref{fig:app_case_l3_map}.}
Although this task may appear similar to a short visual-to-search chain at first glance, it is substantially harder in practice and is better categorized as Level~3. 
The model must first localize the correct region on the map, then reason about a spatial relation relative to a reference entity, then transfer that hypothesis to a different image to extract an auxiliary road cue, and only after that perform external verification. 
This cross-image grounding plus retrieval dependency makes the workflow more than a simple handoff; We therefore treat it as a Level~3 synergistic case rather than a standard Level~2 example.

\vspace{1mm}

\begin{center}
    \includegraphics[width=0.88\linewidth]{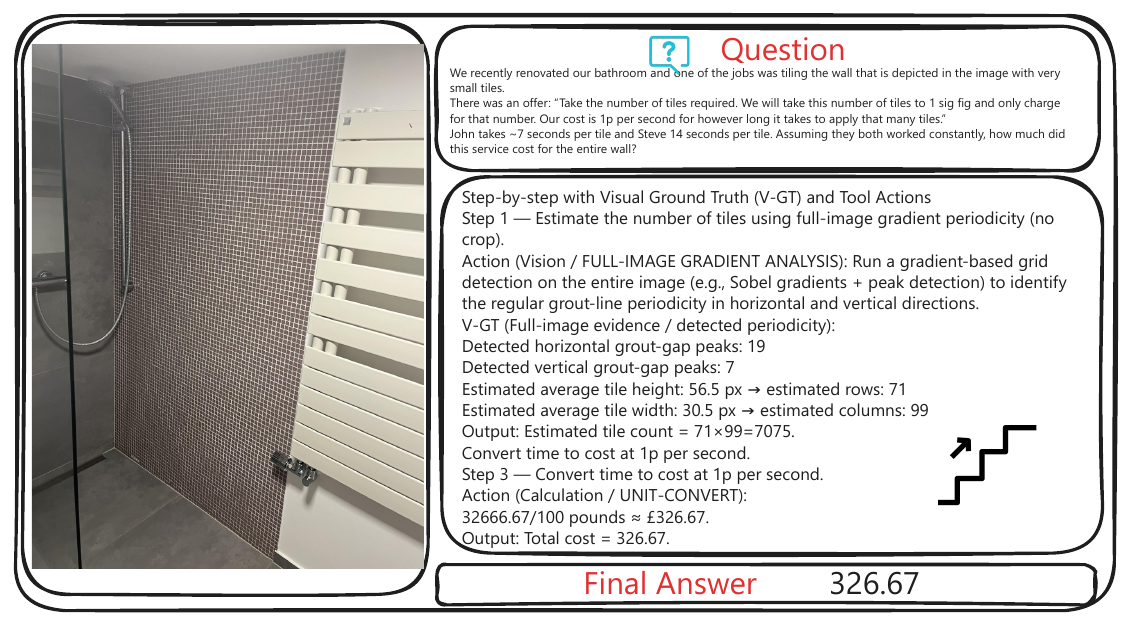}
    \captionof{figure}{\textbf{Level 3 Case: Full-Image Structural Analysis for Dense Tile Estimation.} The agent must infer a latent repetitive grid pattern from the full image, estimate the tile layout, and convert that estimate into a downstream numerical answer.}
    \label{fig:app_case_l3_tile}
\end{center}

\noindent\textbf{Figure~\ref{fig:app_case_l3_tile}.}
This case belongs to Level~3 because the decisive signal is distributed globally across the image rather than concentrated in one local patch. 
The model must reason over large-scale structural regularity, estimate latent periodicity, and then map that estimate to a quantitative answer. 
Such full-image structural analysis goes beyond the crop-and-read regime of Levels~1--2 and instead reflects the advanced CV-analysis branch of Level~3 described in Sec.~\ref{sec:task_setup}.

\vspace{1mm}

\begin{center}
    \includegraphics[width=0.8\linewidth]{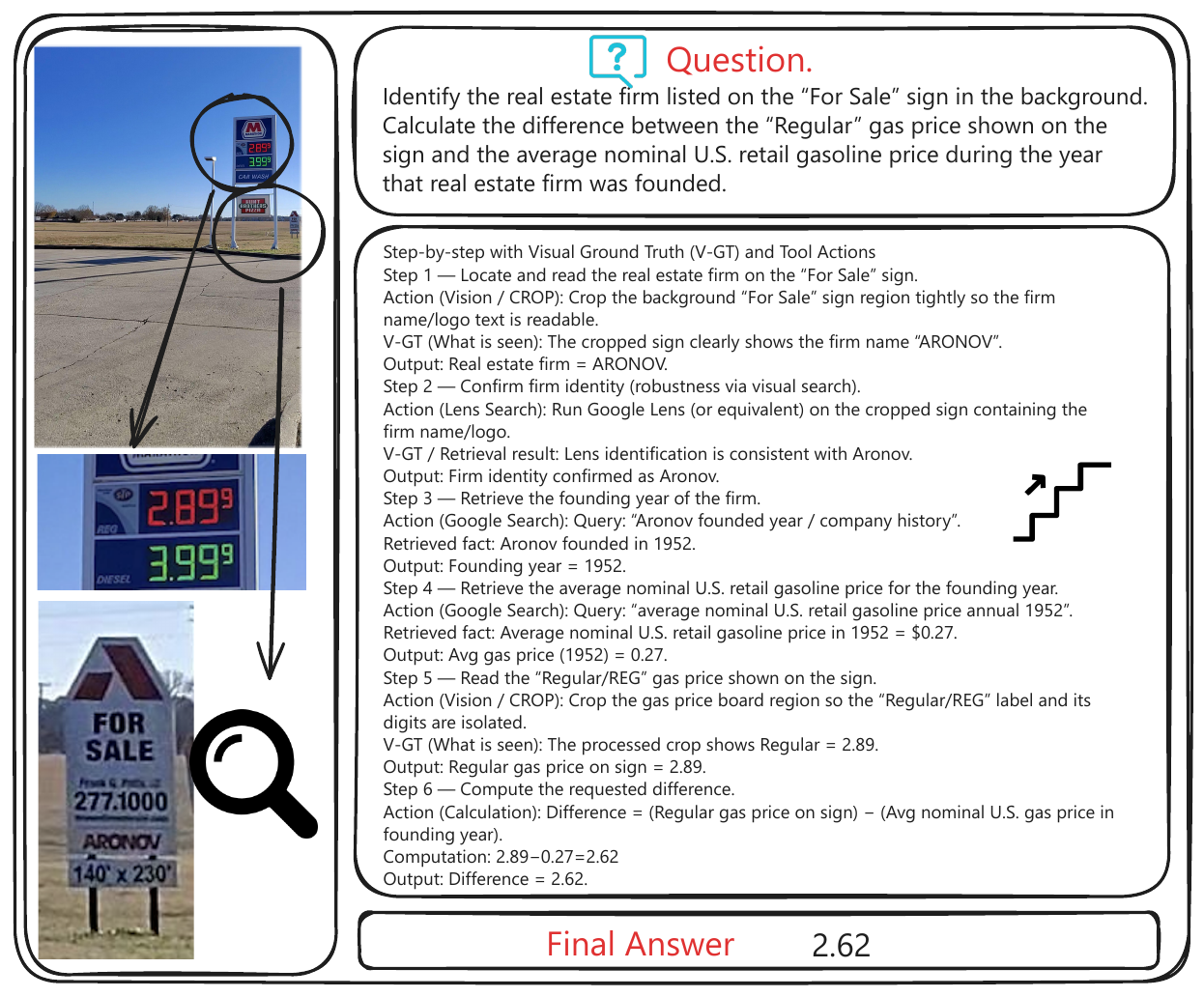}
    \captionof{figure}{\textbf{Level 3 Case: Cross-Domain Multi-Hop Verification.}
    The agent isolates a weak background logo, uses visual search to identify candidate entities, performs multi-hop web retrieval to obtain the required external fact, and then cross-validates this knowledge against another localized cue before computing the final answer.}
    \label{fig:app_case_l3_gas}
\end{center}

\noindent\textbf{Figure~\ref{fig:app_case_l3_gas}.}
This case is a prototypical Level~3 synergistic workflow. 
The initial visual cue is weak and ambiguous, so a single crop is insufficient. 
Instead, the model must propose candidate entities, retrieve external evidence through multiple search hops, and repeatedly cross-check the retrieved knowledge against visual observations before it can finalize the answer. 
Neither isolated visual manipulation nor blind web search can solve this case in isolation.

\vspace{1mm}

\begin{center}
    \includegraphics[width=0.98\linewidth]{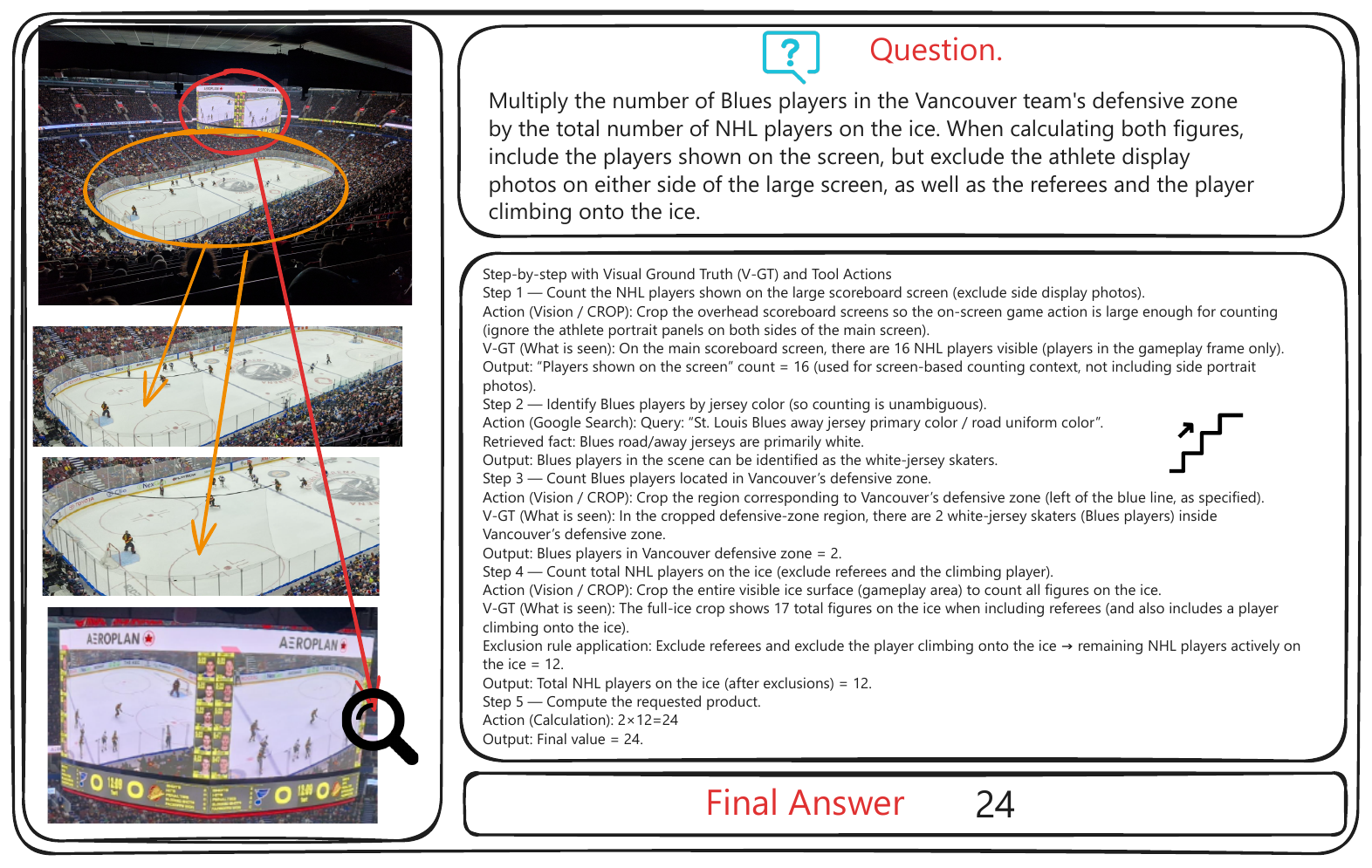}
    \captionof{figure}{\textbf{Level 3 Case: Multi-Step Counting with External Knowledge.}
    The agent coordinates several targeted crops with an external search to identify team jersey colors, and then applies exclusion logic to reach the final count.}
    \label{fig:app_case_l3_hockey}
\end{center}

\noindent\textbf{Figure~\ref{fig:app_case_l3_hockey}.}
This task is also clearly Level~3 because the answer depends on several mutually dependent subproblems: local evidence extraction from different image regions, external identification of team colors, and a final exclusion process that must stay consistent with all previously recovered cues. 
The solution is therefore not a simple chain of operations, but an intertwined reasoning process in which visual observations and retrieved knowledge jointly constrain the answer.
\section{Error Taxonomy}
\label{app:error_taxonomy}

To make the failure analysis in Figure~\ref{fig:error_analysis} reproducible, we define the seven failure modes used in our manual annotation. Each failed trajectory is assigned one dominant label corresponding to the earliest or most consequential bottleneck that prevents successful task completion.

\paragraph{1. Missing search tools.}
The task requires explicit open-web retrieval, but the agent never invokes the search tools at the step where external knowledge is necessary. This includes trajectories that rely only on passive visual inspection or local image manipulation when the answer fundamentally depends on non-visual information.

\paragraph{2. Bad search query.}
The agent does invoke search, but the submitted query is ineffective: it targets the wrong entity, omits the critical visual cue, is too vague to retrieve the intended evidence, or drifts to an unrelated attribute. This category captures failures in transforming localized visual evidence into a useful retrieval query.

\paragraph{3. Unfaithful visual tool use.}
The agent chooses to use a visual tool, but the produced artifact does not faithfully expose the required evidence. Typical examples include cropping the wrong region, rotating in the wrong direction, over-enhancing into unreadability, or otherwise generating an intermediate image that fails the V-axis verification.

\paragraph{4. Missing visual tool use.}
The task requires an explicit visual manipulation step, but the agent never performs it. This includes cases where the model answers directly from the raw image, prematurely switches to search, or repeatedly reasons in natural language without taking the necessary visual action.

\paragraph{5. Overthinking Collapse.}
The agent enters a redundant exploration loop after the necessary evidence could already have been obtained. Typical symptoms include repeated near-duplicate crops, unnecessary additional searches, repeated verification attempts, or excessive trial-and-error that derails the trajectory and wastes interaction budget.

\paragraph{6. Tool-Misexecution.}
The trajectory fails because of interface-level execution mistakes rather than reasoning alone. Examples include malformed code, invalid tool arguments, runtime errors, missing file saves, or other failures that prevent the intended tool action from being executed correctly.

\paragraph{7. PostVisual-Perception-Deficit.}
The agent successfully produces a relevant intermediate artifact, but still fails to correctly perceive or read the required evidence from that artifact. In other words, the bottleneck is no longer whether the correct region was surfaced, but whether the model can interpret the surfaced visual cue itself.

\paragraph{Annotation rule.}
When multiple problems appear in the same failed trajectory, we assign the dominant label according to the earliest or most consequential bottleneck. For example, if the model fails to crop the relevant region and later also issues a poor search query, the case is labeled as \textit{Unfaithful visual tool use} rather than \textit{Bad search query}.

\subsection{System Prompts for Gen and Atm Modes}
\label{app:system_prompts}

To improve reproducibility, we provide the exact system-prompt templates used in the released evaluation harness. 
The benchmark supports two interaction modes: \textbf{Gen} mode, where the model writes sandboxed Python code for visual processing while invoking search tools via function calling; and \textbf{Atm} mode, where the model interacts through a fully atomic function-calling interface. 
These prompts are not merely stylistic wrappers: they define image indexing, output formatting, action/answer separation, and the operational constraints under which all models are evaluated.

\paragraph{Gen mode (ReAct-style code + tool use).}
The Gen-mode prompt combines ReAct-style stepwise reasoning with two execution channels: (i) search tools through function calling, and (ii) local code execution through \texttt{<code>} blocks. 
It explicitly defines image indexing, processed-image naming conventions, and the rule that an agent may not produce a final answer in the same turn as a tool action.

\begin{tcolorbox}[
    enhanced,
    breakable,
    colframe=black!70,
    colback=yellow!5,
    boxrule=1pt,
    arc=4mm,
    left=2mm,
    right=2mm,
    top=1mm,
    bottom=1mm,
    title={Released system prompt: Gen / ReAct mode}
]
\small
\begin{verbatim}
You are a multimodal reasoning agent that solves visual questions
step by step.

You have access to:
1. **Search tools** (via function calling): google_search, google_lens_search, fetch_webpage.
2. **Code execution**: Write Python code in <code> blocks for
   image manipulation and analysis

## Image Management
- Images are tracked by index: Image 0 is the original input,
  Images 1, 2, ... are processed results
- Image N corresponds to transformed_image_N.png
  (e.g., Image 1 = transformed_image_1.png)
- After your code runs, new images will be shown with their index
  (e.g., "[Image 1: transformed_image_1.png]")
- You can reference any image by its index when using search tools

## Workflow (ReAct Pattern)

For each step:
1. **Think**: Analyze what you know and what you need
2. **Act**: Use search tools OR write code as needed
3. **Observe**: Review results
4. **Repeat** until you have enough information
5. **Answer**: Provide your final answer

## Response Format

Use these XML blocks as needed (all are OPTIONAL):

<think>
Your reasoning process. Analyze the image, plan your approach,
interpret tool results.
</think>

<code>
Python code for image processing.

Available paths (via environment variables):
- os.environ['ORIGINAL_IMAGE_PATH']: Path to the original input image
  (Image 0)
- os.environ['PROCESSED_IMAGE_SAVE_PATH']: Directory to save
  processed images

Naming convention for saved images:
- Save as: transformed_image_1.png, transformed_image_2.png, etc.
  (starting from 1)
- Full path:
  os.path.join(os.environ['PROCESSED_IMAGE_SAVE_PATH'],
  'transformed_image_1.png')
- Image N corresponds to transformed_image_N.png

You can read any previously saved image from the output directory,
including downloaded images (downloaded_image_N.png).
Libraries available: PIL, cv2, numpy, matplotlib, scipy
Use print() to output values. Do NOT use display() or plt.show().
</code>

<answer>
Your final answer. Only include when you have enough information.
</answer>

## Critical Rules

1. **Do NOT combine action and answer in the same turn**:
   - If you use <code> or call a search tool, do NOT include
     <answer> in the same response
   - Wait for the results before providing your answer
   - <answer> should only appear when you are ready to give the
     final answer with NO more actions needed

2. **Image feedback**: After your code runs, you will automatically
   receive:
   - The stdout/stderr output
   - New images with their indices
   - All newly generated images displayed directly

3. **Using specific images with search tools**:
   - Use google_lens_search with "image_path" parameter to search a
     specific image
   - Example: {"image_path": "transformed_image_1.png"} to search
     Image 1
   - Or use "image_ref": "original" for Image 0, "current" for the
     latest image

## Important

- Search tools are called via function calling, NOT in <code>
- Code in <code> blocks will be executed locally
- Think step by step in <think>
- Only provide <answer> when confident and after observing all
  results
\end{verbatim}
\end{tcolorbox}

\paragraph{Atm mode (atomic function calling).}
The Atm-mode prompt removes free-form code writing and instead exposes a fixed tool interface. 
Its main purpose is to normalize image indexing, tool arguments, and the action protocol across providers. 
In the released public prompt, the retrieval interface exposes three active search tools, while \texttt{download\_image} is documented but disabled.

\begin{tcolorbox}[
    enhanced,
    breakable,
    colframe=black!70,
    colback=yellow!5,
    boxrule=1pt,
    arc=4mm,
    left=2mm,
    right=2mm,
    top=1mm,
    bottom=1mm,
    title={Released system prompt: Atm / atomic mode}
]
\small
\begin{verbatim}
You are a multimodal reasoning agent with access to image
manipulation and web search tools.

## Image Management
- Images are tracked by index: Image 0 is the original input,
  Images 1, 2, ... are processed results
- Image N corresponds to transformed_image_N.png
  (e.g., Image 1 = transformed_image_1.png)
- Each tool operation creates a NEW image with a new index
- You must specify which image to operate on using `image_index`
  parameter
- After each operation, you'll see the new image and its index

## Image Tools (function calling)
All tools require `image_index` to specify which image to operate on.

Geometric transformations:
- crop(image_index, bbox_2d, zoom_scale?, label?) - Crop a region
  using normalized coordinates [x1,y1,x2,y2] in 0-1000 scale
- rotate(image_index, angle, expand?, label?) - Rotate the image
- flip(image_index, direction?, label?) - Flip/mirror the image
  (horizontal/vertical/both)
- resize(image_index, width?, height?, scale?, label?) - Resize the
  image

Enhancement/filtering:
- enhance(image_index, brightness?, contrast?, sharpness?, label?) -
  Adjust brightness/contrast/sharpness (1.0=no change)
- grayscale(image_index, label?) - Convert to grayscale
- autocontrast(image_index, cutoff?, label?) - Automatic contrast
  adjustment
- blur(image_index, radius?, label?) - Apply Gaussian blur
- sharpen(image_index, label?) - Apply sharpening filter
- denoise(image_index, strength?, label?) - Remove noise
- edge_detect(image_index, method?, label?) - Detect edges
  (canny/sobel/simple)
- invert(image_index, label?) - Invert colors (negative)
- equalize(image_index, label?) - Equalize histogram
- threshold(image_index, value?, mode?, label?) - Convert to binary

## Coordinate System
- bbox_2d uses normalized coordinates: [x1, y1, x2, y2] where each
  value is 0-1000
- (0, 0) is top-left, (1000, 1000) is bottom-right
- Example: [250, 250, 750, 750] crops the center 50% of the image

## Web Search Tools
- google_search(query, gl?, hl?) - Text-based web search
- google_lens_search(image_index?) - Reverse image search on
  specified image (default: 0 = original)
- fetch_webpage(url, max_chars?) - Fetch webpage content
 - download_image(url) - Download an image from URL
  (max 5 per task).

## Workflow
1. Analyze the image (Image 0) and question
2. Use tools as needed, always specifying image_index
3. After each tool, you'll see the result and new image index
4. Continue until you have enough information
5. Provide your final answer with this REQUIRED format:
   <answer>YOUR_FINAL_ANSWER</answer>
\end{verbatim}
\end{tcolorbox}

\paragraph{What these prompts standardize.}
The two prompts normalize four aspects that are essential for fair evaluation:
(i) \textbf{image-state bookkeeping}, by enforcing a shared image-index protocol;
(ii) \textbf{action/answer separation}, by forbidding tool use and final answer generation in the same turn;
(iii) \textbf{artifact naming and logging}, especially in Gen mode via \texttt{transformed\_image\_N.png}; and
(iv) \textbf{tool exposure}, by making the available visual and retrieval operations explicit to the agent.
\end{document}